\pdfoutput=1

\documentclass[11pt]{article}

\usepackage[final]{acl}

\usepackage{times}
\usepackage{latexsym}

\usepackage[T1]{fontenc}

\usepackage[utf8]{inputenc}

\usepackage{microtype}

\usepackage{inconsolata}

\usepackage{graphicx}
\definecolor{cvprblue}{rgb}{0.21,0.49,0.74}
\definecolor{IllinoisOrange}{HTML}{FF5E00}
\definecolor{IllinoisBlue}{HTML}{406080}
\definecolor{VTMaroon}{HTML}{630031}
\definecolor{LightMaroon}{HTML}{922050}

\usepackage{enumitem}
\usepackage{amsmath}
\usepackage{amssymb}
\usepackage{booktabs}
\usepackage{xspace}
\usepackage{multirow,tabularx}
\usepackage{caption}
\usepackage{bm}
\usepackage{colortbl}
\usepackage{pgfplots}
\usetikzlibrary{plotmarks}
\usepackage{tcolorbox}
\usepackage{pifont}
\usepackage{csquotes}
\usepackage{fancyvrb}
\usepackage{tabularx}
\definecolor{mygraylite}{gray}{.94}
\definecolor{mygray}{gray}{.89}
\definecolor{green}{HTML}{009400}
\definecolor{darkergreen}{RGB}{21, 152, 56}
\definecolor{amber}{rgb}{1.0, 0.75, 0.0}
\definecolor{darkseagreen}{rgb}{0.56, 0.74, 0.56}
\definecolor{babyblueeyes}{rgb}{0.63, 0.79, 0.95}
\definecolor{burntsienna}{rgb}{0.91, 0.45, 0.32}
\definecolor{myviolet}{rgb}{0.57, 0.51, 0.9}
\definecolor{amethyst}{rgb}{0.6, 0.4, 0.8}
\definecolor{ashgrey}{rgb}{0.7, 0.75, 0.71}
\definecolor{battleshipgrey}{rgb}{0.35, 0.35, 0.35}
\definecolor{asparagus}{rgb}{0.53, 0.66, 0.42}
\definecolor{maroon}{RGB}{150, 0, 0}
\newcommand*{\eg}{\emph{e.g.}}
\newcommand*{\ie}{\emph{i.e.}}
\usepackage{nicematrix}
\newcommand*{\etc}{\emph{etc}}
\usepackage{fontawesome}

\DeclareMathOperator*{\argmax}{arg\,max}

\definecolor{mmPurple}{RGB}{106, 58, 145}
\newcommand{\modelname}{\textcolor{mmPurple}{\textbf{MMPlanner}}\@\xspace}
\newcommand{\modelnamenc}{{{MMPlanner}}\@\xspace}
\newcommand*{\planscore}{{{{T-PlanScore}}}\@\xspace}
\newcommand*{\imagetotextrelevance}{{{{CA-Score}}}\@\xspace}
\newcommand*{\visualsequenceordering}{{{{VS-Ordering}}}\@\xspace}

\usepackage{fancyvrb,newverbs,xcolor}
\definecolor{cverbbg}{rgb}{0.94, 0.97, 1.0}
\newenvironment{lcverbatim}
 {\VerbatimEnvironment
  \begin{Verbatim}[commandchars=\\\{\}]}
 {\end{Verbatim}}

\newsavebox{\cverbbox}
\newenvironment{coloredlcverbatim}
 {\VerbatimEnvironment
  \begin{lrbox}{\cverbbox}
  \begin{footnotesize}
  \begin{minipage}{0.98\columnwidth}
  \begin{lcverbatim}}
 {\end{lcverbatim}
  \end{minipage}
  \end{footnotesize}
  \end{lrbox}
  \begin{center}
  \colorbox{cverbbg}{\usebox{\cverbbox}}
  \end{center}}

\usepackage[capitalize]{cleveref}
\crefname{section}{Sec.}{Secs.}
\Crefname{section}{Section}{Sections}
\Crefname{table}{Table}{Tables}
\crefname{table}{Tab.}{Tabs.}

\title{
    \textbf{\LARGE \modelname: Zero-Shot  Multimodal Procedural Planning with Chain-of-Thought Object State Reasoning}
}

\author{Afrina Tabassum\\
  Amazon \\
  \texttt{afrinat@amazon.edu} \\\And
  Bin Guo \\
  Alexa, Amazon \\
  \texttt{guobg@amazon.com} \\\And
  Xiyao Ma \\
  Alexa, Amazon \\
  \texttt{maxiya@amazon.com} \\\AND
  Hoda Eldardiry\\
  Virginia Tech \\
  \texttt{hdardiry@vt.edu} \\\And
  Ismini Lourentzou \\
  University of Illinois Urbana-Champaign \\
  \texttt{lourent2@illinois.edu} 
}

\begin{document}

\maketitle
\begin{abstract}
Multimodal Procedural Planning (MPP) aims to generate step-by-step instructions that combine text and images, with the central challenge of preserving object-state consistency across modalities while producing informative plans. Existing approaches often leverage large language models (LLMs) to refine textual steps; however, visual object-state alignment and systematic evaluation are largely underexplored.
We present \modelname, a zero-shot MPP framework that introduces Object State Reasoning Chain-of-Thought (OSR-CoT) prompting to explicitly model object-state transitions and generate accurate multimodal plans. To assess plan quality, we design LLM-as-a-judge protocols for planning accuracy and cross-modal alignment, and further propose a visual step-reordering task to measure temporal coherence.
Experiments on \textsc{RecipePlan} and \textsc{WikiPlan} show that \modelnamenc achieves state-of-the-art performance, improving textual planning by $+6.8\%$, cross-modal alignment by $+11.9\%$, and visual step ordering by $+26.7\%$.
\vspace{0.1cm}
\color{mmPurple}{\faGlobe}~\href{https://plan-lab.github.io/mmplanner}{https://plan-lab.github.io/mmplanner}
\end{abstract}
\begin{figure*}[t!]
  \centering
  \includegraphics[width=0.99\linewidth]{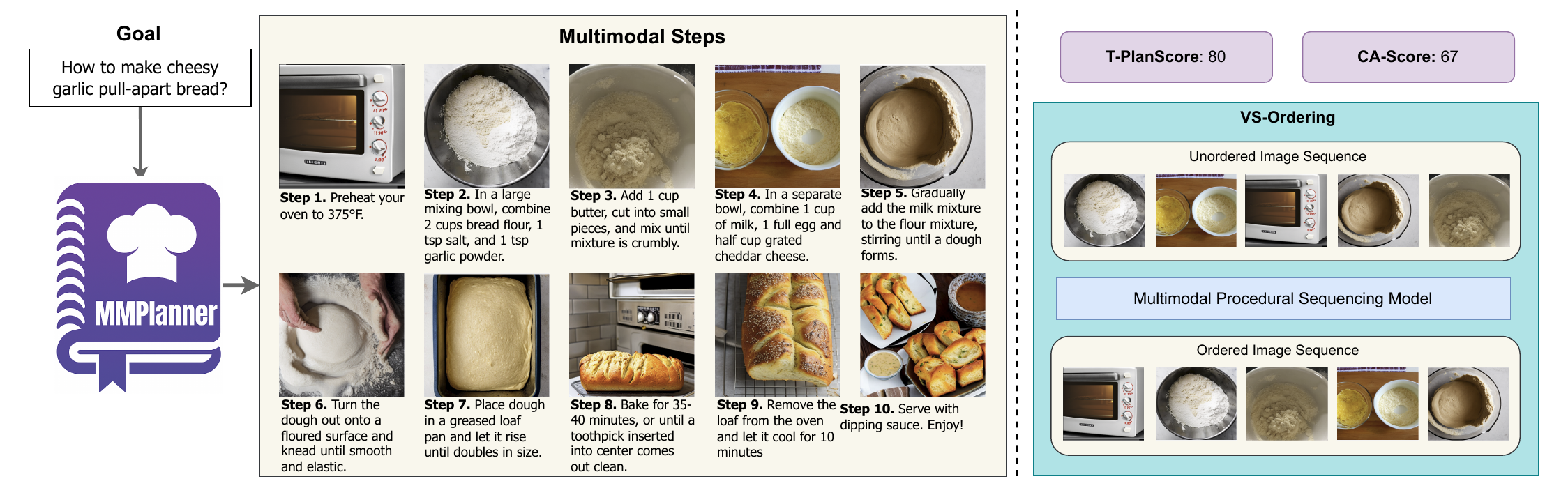}
  \vspace{-0.3cm}
  \caption{\textbf{Multimodal Procedural Planning.} Left: \modelname, processes overall goals to produce comprehensive step-by-step textual and visual plans. Right: Our proposed evaluation assesses the planning accuracy of the textual plan, the cross-modal alignment between visual and textual steps, and the temporal coherence of the visual steps.}
  \label{fig:teaser}
  \vspace{-0.3cm}
\end{figure*}

\section{Introduction}
Procedural planning involves generating a sequence of steps to accomplish a goal~\citep{lyu2021goal}, \eg, \textit{baking a cake} or \textit{assembling a bookshelf}. Domains such as robotics~\citep{kovalchuk2021verifying, zhao2023large}, reasoning systems~\citep{chen2017survey, wei2022chain}, \etc., rely on effective procedural planning. Consequently, the field has received growing attention, driven by the recent advancements in LLMs~\citep{liu2023improvedllava, zhu2023minigpt}. Existing works utilize task-specific concept knowledge~\citep{sun-etal-2023-incorporating}, knowledge from LLMs~\citep{yuan-etal-2023-distilling}, or multimodal input~\citep{zhou-etal-2023-non, wang-etal-2023-multimedia}, and generate linear~\citep{wang-etal-2023-multimedia, yuan-etal-2023-distilling, sun-etal-2023-incorporating} or non-linear~\citep{zhou-etal-2023-non} textual procedural plans. 
However, text-only instructions often lack the visual clarity and specificity required for complex tasks, limiting understanding, accessibility, and engagement. Multimodal Procedural Planning (MPP) addresses these limitations by jointly generating textual step instructions with corresponding step images, yielding more precise and accessible procedural knowledge.

A central challenge in MPP is generating step visuals that accurately reflect object state transitions. These transitions can be explicit, when the change is clearly described in the current textual step, or implicit, when it must be inferred from prior steps or broader context. For instance, in Figure~\ref{fig:teaser}, step 2 explicitly describes mixing ingredients, so the corresponding image should depict a bowl containing the mixed dry ingredients. 
Step 3, in contrast, involves adding butter, and the image must implicitly incorporate the existing mixture from Step 2. In this case, the visual should show butter being added into the bowl of mixed dry ingredients, even though ingredients are not restated in the step text.

Another important challenge involves the evaluation of multimodal plans, particularly in determining whether the generated steps successfully accomplish the intended task. Prior work has primarily measured semantic similarity between generated and reference textual plans~\citep{wang-etal-2023-multimedia, yuan-etal-2023-distilling}, which cannot effectively verify true task completion. Furthermore, text-based semantic metrics overlook critical dimensions such as visual–text alignment, temporal coherence, and the informativeness of visual steps. As a result, evaluation of multimodal plans still relies heavily on human judgment~\citep{lu2023multimodal, wang-etal-2023-multimedia}, which is labor-intensive, difficult to scale, and prone to inconsistency across annotators.

To address these challenges, we introduce \textbf{\underline{M}}ulti\textbf{\underline{M}}odal \textbf{\underline{Planner}} (\textbf{\modelname}), a zero-shot framework for generating consistent multimodal plans that capture both explicit and implicit object state changes across visual steps. \modelnamenc leverages Chain-of-Thought ({CoT}) prompting with \underline{\textbf{O}}bject \underline{\textbf{S}}tate \underline{\textbf{R}}easoning (\textbf{OSR-CoT}) to guide the model in reasoning about object transitions across steps.
To the best of our knowledge, \modelnamenc is the first MPP approach to generate zero-shot multimodal plans that jointly model implicit and explicit state changes through prompted reasoning, without requiring task-specific training.

To enable scalable and automatic evaluation of MPP, we introduce a set of customized multimodal LLM-based evaluators and propose a comprehensive evaluation framework comprising: 
(1) \underline{\textbf{T}}extual-\textbf{Plan} \textbf{Score} (\textbf{\planscore}), which measures the planning accuracy of the generated textual plan by assessing the alignment between the goal and the generated textual steps;
(2) \underline{\textbf{C}}ross-modal \underline{\textbf{A}}lignment \textbf{Score} (\textbf{\imagetotextrelevance}),
which evaluates the relevance between generated step images and corresponding textual steps; and
(3) \underline{\textbf{V}}isual \underline{\textbf{S}}tep \textbf{Ordering} (\textbf{\visualsequenceordering}), a task that assesses the informativeness and temporal coherence of the visual plan by recovering the correct step order from shuffled images.
The contributions of our work are: 
\begin{itemize}[topsep=0pt, partopsep=0pt, itemsep=0pt, parsep=0pt]
\item We introduce \textbf{\modelname}, a zero-shot MPP method that generates coherent multimodal plans reflecting object state changes in visual plan sequences. We empirically validate \modelnamenc on two benchmark datasets, achieving improvements of up to $6.8\%$ in textual plan quality, $11.9\%$ in cross-modal alignment, and $26.7\%$ in visual step ordering accuracy.

\item \modelnamenc incorporates background context from previous task steps through an \textbf{Object State Reasoning Chain-of-Thought (OSR-CoT)} prompting strategy, enabling explicit modeling of evolving object states across steps, and reducing inference time by $\sim$46.25\% compared to SoTA MPP baselines.

\item We propose a \textbf{reference-free evaluation} framework to assess planning accuracy, cross-modal alignment, temporal coherence, and visual informativeness of generated plans, achieving stronger correlation with human judgments than prior cross-modal metrics ($\rho\!=\!0.57$ vs.\ $0.37$ for CLIPScore) while reducing step-level evaluation time by $\sim$66\% (90s $\to$ 30s). For textual planning, our automated evaluation requires only $\sim$0.7s per plan compared to $\sim$5mins for human assessment, enabling large-scale evaluation.

\end{itemize}

\begin{figure*}[t!]
  \centering
  \includegraphics[width=0.99\textwidth]{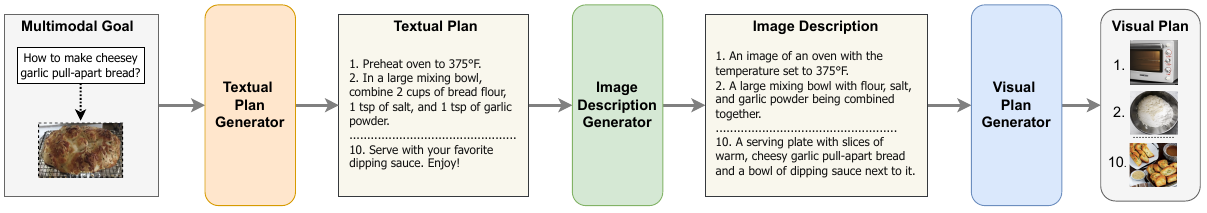}
  \vspace{-0.3cm}
  \caption{\textbf{\modelname Overview.} Given a goal instruction, \modelnamenc first generates a corresponding visual goal. Then, the Textual Plan Generator produces textual plans aligned with this multimodal goal. Each step is passed to the Image Description Generator, which produces detailed visual descriptions capturing explicit and implicit object state changes. Finally, the Visual Plan Generator uses these descriptions to create step-by-step images, resulting in a comprehensive multimodal plan that maintains consistency across both textual and visual steps.}
  \label{fig:overview}
\end{figure*}

\section{Related Work}
\noindent\textbf{Procedural Planning.} Procedural planning methods fall into two categories: selection-based and generation-based. Selection-based approaches~\citep{zhao2023large, lu2022neuro, song2023llm, wu2022understanding, zhou2022show, ashutosh2023video} rely on predefined candidates, limiting generalization to unseen scenarios. Generation-based methods, powered by LLMs~\citep{zhu2023minigpt, ouyang2022training}, focus on generating textual plans~\citep{wang-etal-2023-multimedia, sun-etal-2023-incorporating}. Recently, TIP~\citep{lu2023multimodal} generates multimodal plans by prompting an LLM and image generation model twice sequentially, increasing the inference time. In summary, existing methods often fail to accurately reflect changes in object states throughout the steps. 

Although recent work explores tracking state changes in videos~\cite{niu2024schema} and leverages large-scale datasets~\cite {souvcek2024showhowto}, these approaches typically assume access to full video sequences~\cite{niu2024schema} and struggle to maintain state consistency across frames~\cite{souvcek2024showhowto}. In addition, Statler~\citep{yoneda2023statler} focuses on maintaining object states for embodied robotic reasoning with low-level, fine-grained actions in closed environments. In contrast, our work generates visual plans from scratch in a zero-shot setting, given only a goal, and emphasizes high-level, interpretable steps aimed at human-centric multimodal procedural planning.

Beyond procedural planning, prior work has explored cross-modal coherence and multimodal discourse, focusing on temporal and narrative consistency across modalities \cite{alikhani-etal-2019-cite,inan-etal-2021-cosmic}. Such studies show that modeling discourse relations and coherence cues can strengthen alignment and narrative flow. Our work is complementary, and future extensions could incorporate discourse-aware prompting to further enhance temporal and narrative consistency in visual plans.

\noindent\textbf{Multimodal Plan Evaluation.} 
Procedural plans can be evaluated manually or automatically. Human evaluations (\eg, crowdsourcing) can be time-intensive and error-prone, while automatic metrics such as 
WMD~\citep{kusner2015word}, Sentence-BERT~\citep{reimers2019sentence}, \etc., though scalable, fall short in assessing temporal relationships and completeness of textual plans. 
For evaluating multimodal plans, prior work measures similarity between textual plans and captions from visual plans, encountering similar limitations~\cite{lu2023multimodal}.
To the best of our knowledge, there is currently a lack of automatic frameworks for evaluating the quality of multimodal plans. To address this gap, we propose \planscore for task completion and \imagetotextrelevance for cross-modal alignment evaluation. Additionally, we introduce a visual reordering task to assess temporal coherence.

\section{Method}
The goal of MPP is to generate a sequence of steps, each with a textual and visual component, that together achieve a high-level task goal. 
Given a high-level goal instruction $\mathcal{G}_t$ that outlines the task, the objective is to generate a sequence of low-level steps $\mathcal{S}\!=\!\{s_1, s_2, \ldots, s_n\}$, where $n$ denotes the number of steps. Each step $s_i$ comprises a textual description $t_i$ and a corresponding step image $v_i$, denoted as $(t_i, v_i)$. The step-wise textual and visual plans are denoted as  $\mathcal{T}\!=\!\{t_1, t_2, \ldots, t_n\}$ and $\mathcal{V}\!=\!\{v_1, v_2, \ldots, v_n\}$, respectively.

\subsection{\textbf{\modelnamenc} Overview}
We introduce {\modelnamenc}, a method for generating a multimodal plan $\mathcal{S}\!=\!(\mathcal{T}, \mathcal{V})$  from a given task goal $\mathcal{G}_t$, consisting of (1) a {Textual Plan Generator} that produces a sequence of textual steps $\mathcal{T}$ from goal $\mathcal{G}_t$, (2) an {Image Description Generator} that produces detailed image descriptions $\mathcal{D}$ from textual steps $\mathcal{T}$, capturing both explicit and implicit object state changes, and (3) a {Visual Plan Generator} that generates visual plans $\mathcal{V}$ from descriptions $\mathcal{D}$. An overview is shown in Figure~\ref{fig:overview}.

\subsection{Textual Plan Generator}\label{subsec:textual_plan_generator}
Recent advancements in LLMs~\citep{taori2023stanford} have facilitated the generation of step-by-step textual plans from a high-level goal~\citep{lu2023multimodal, wang-etal-2023-multimedia, sun-etal-2023-incorporating, yuan-etal-2023-distilling}. 
However, text-only goal instructions $\mathcal{G}_t$ often under-specify the task and omit visual cues such as object appearances or spatial configurations that are crucial for accurate plan generation. These missing cues can lead to ambiguous or incomplete plans, especially in tasks requiring implicit state reasoning.
To address this, we enhance task comprehension by extending goal instructions $\mathcal{G}_t$ to include a corresponding visual goal $\mathcal{G}_v$, generated using Stable Diffusion~\citep{rombach2022high}. The resulting multimodal goal is denoted as $\mathcal{G}\!=\!(\mathcal{G}_t, \mathcal{G}_v)$, where $\mathcal{G}_t$ and $\mathcal{G}_v$ represent textual and visual goals, respectively. 
We then utilize a VLM~\citep{liu2023improvedllava} to generate a step-by-step textual plan $\mathcal{T}$ from $\mathcal{G}$. 

\subsection{Image Description Generator}\label{subsec:image_description_generator}
\begin{figure}[t!]
\centering
\vspace{-0.1cm}
\includegraphics[width=0.97\columnwidth]{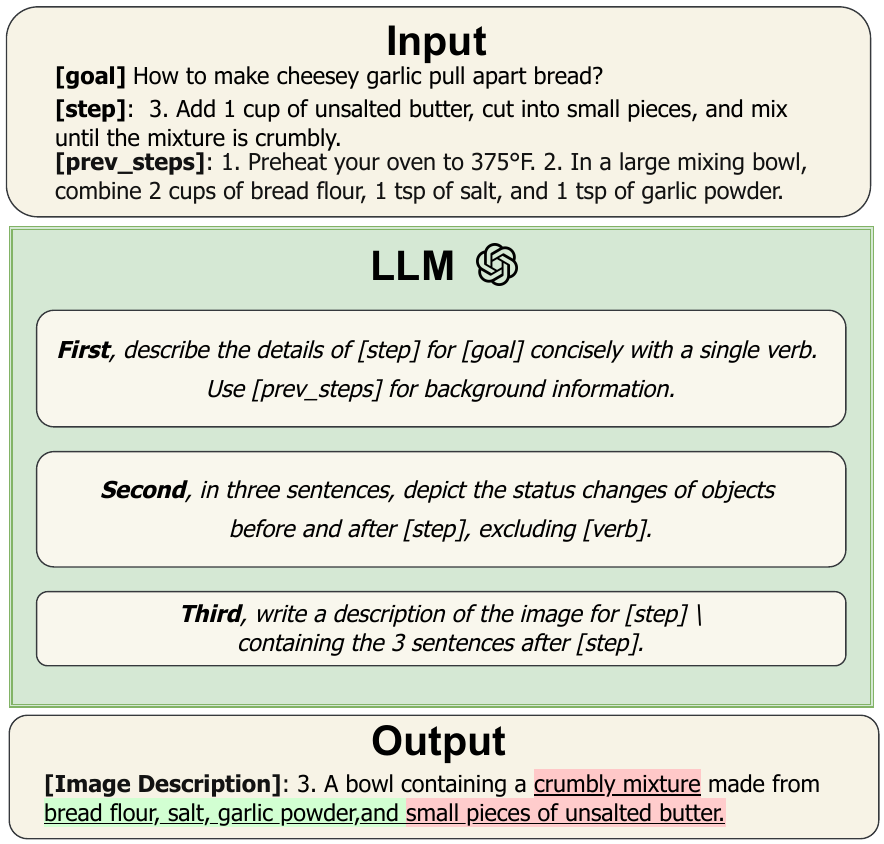}
  \vspace{-0.3cm}
  \caption{\textbf{Image Description Generator}. Our proposed OSR-CoT prompts the LLM to include both explicit (red) and implicit (green) state changes of the associated objects in the generated image description.}
    \label{fig:image_description_generator}
    \vspace{-0.2cm}
\end{figure}

While Text-to-Image (T2I) generation models can produce images from the information explicitly present in the textual descriptions~\citep{rombach2022high}, they struggle to interpret implicit state changes from textual steps.
For example, when prompted with the step instructions \enquote{\textit{3. Add 1 cup of unsalted butter, cut into small pieces, and mix until the mixture is
crumbly}}, Stable Diffusion produces an image of butter, unable to infer the concept of a \enquote{mixture} due to the lack of explicit information present in the prompt. 
To address this limitation, following recent works~\citep{niu2024schema,menon2022visual}, we employ an LLM~\citep{brown2020language} to generate image descriptions from textual plans, leveraging inherent commonsense knowledge. Specifically, we feed the LLM with each textual step along with the overall goal and previous steps to offer additional context and enable it to infer implicit object state changes. 
A corresponding simple prompt for generating image descriptions is as follows: 
\vspace{-15pt}

\begin{coloredlcverbatim}
Prompt: 
In the process of [goal], current step is [step].
The previous steps are [prev_steps]. Describe an 
image containing the items involved in the  
current step, after completing the current step. 
Focus on the items and their physical states.
Answer: <Image Description>
\end{coloredlcverbatim}
\vspace{-2pt}
However, with this prompt, the model neglects key details like texture and often struggles to contextualize previous steps as background information, thereby tending to include extraneous details from previous steps in the image description of the current step. For instance, for the input [step] illustrated in Figure~\ref{fig:image_description_generator}, the above prompt generates: 

\noindent \enquote{\small \textit{The image shows a bowl of bread flour mixture with small pieces of unsalted butter. The flour is white and powdery, while the salt and garlic powder are both fine grains}}. 

\noindent Here, the model overlooks the texture \enquote{crumbly mixture} and hallucinates irrelevant details about flour, salt, and garlic powder from previous steps.

\noindent\textbf{Chain-of-Thought Prompting with Object State
Reasoning (OSR-CoT):} To address these challenges, inspired by ~\cite{niu2024schema}, we introduce OSR-CoT, a Chain-of-Thought (CoT) prompting strategy designed to reduce hallucinations by guiding the model through a stepwise reasoning process. 
First, OSR-CoT prompts the model to describe the current step in detail, incorporating relevant background information from the overall goal and previous steps. 
Next, OSR-CoT instructs the model to reason about object state changes before and after the current step. Finally, it directs the model to incorporate these state changes into a coherent image description.
As illustrated in Figure~\ref{fig:image_description_generator}, OSR-CoT generates concise image descriptions, including both explicit and implicit object state changes.
Unlike the previously introduced simple prompt, the image description generated by OSR-CoT (Figure~\ref{fig:image_description_generator}) is concise without unnecessary hallucinated information. We denote the sequence of generated image descriptions as $\mathcal{D}\!=\!\{d_1,\ldots,d_i, \ldots, d_n\}$ where $n$ is the total number of steps and $d_i$ is the image description for the $i$-th step.

\subsection{Visual Plan Generation}\label{subsec:visual_plan_generation}
\begin{figure}[t!]
\centering
  \includegraphics[width=0.95\columnwidth]{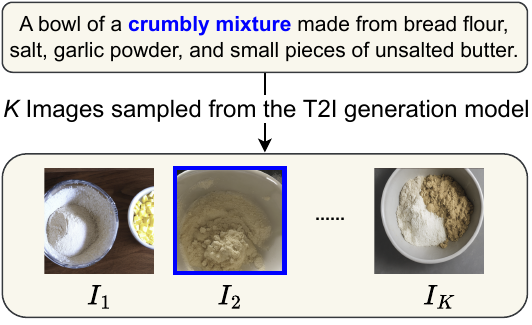}
  \vspace{-0.3cm}
  \caption{\textbf{Cross-modal Step Image Selection} showing multiple images generated with SD given the same image description as prompt for step \enquote{\textit{3. Add 1 cup of unsalted butter, cut into small pieces, and mix until the mixture is crumbly}}. Here, $I_2$ successfully captures \enquote{\textit{crumbly}} and \enquote{\textit{mixture}} textures \textcolor{blue}{(blue bbox)}, whereas $I_1$ and $I_K$ fail to incorporate these fine-grained details.}
  \label{fig:image_generation}
\vspace{-0.2cm}
\end{figure}

Given a step image description $d_i$, we generate a corresponding step image $v_i$ using Stable Diffusion~\citep{rombach2022high}.
Since Stable Diffusion is a stochastic generative model, prompting it multiple times with the same description $d_i$ yields a set of diverse images $\mathcal{I}_i\!=\!\{I_{i1}, \ldots, I_{ik}, \ldots, I_{iK}\}$. Empirically, we observe that while some samples accurately reflect the fine-grained attributes in $d_i$, others may miss key visual details. As shown in Figure~\ref{fig:image_generation}, only $I_2$ captures both the \enquote{crumbly} texture and the \enquote{mixture} mentioned in the description, whereas $I_1$ and $I_K$ fail to depict these elements. To ensure consistency and visual fidelity across steps, we sample multiple candidates and select the one that best aligns with the textual description.

We introduce a cross-modal selection strategy to choose the best image from $\mathcal{I}_i$. Each image sample is assigned a similarity score based on its alignment with the description $d_i$ in the feature space. 
To map $d_i$ and $\mathcal{I}_i$ into a shared feature space, we utilize a pretrained BLIP-2~\citep{li2023blip} feature extractor. In BLIP-2, a Querying Transformer (Q-Former) is trained to bridge the gap between a frozen image encoder and an LLM, where the objective is to generate a visual feature representation that is relevant to the prompt and interpretable by the LLM.
Let $f_{d_i}$ and $f_{ik}$ denote the BLIP-2 feature embeddings of the image description $d_i$ and the $k$-{th} sampled image $I_{ik}$, respectively, where $1\leq k\leq K$. 
We select an image based on cross-modal similarity 
$\argmax_{k} sim(f_{ik}, f_{d_i})$,
where $sim(\cdot,\cdot)$ refers to cosine similarity. This process is repeated for all steps to obtain the final visual plan $\mathcal{V}$ consisting of all generated step images.

\section{MPP Evaluation}
\label{sec:multimodal_evaluation}
LLMs have demonstrated strong performance on complex reasoning tasks, motivating their use as automated evaluators that often surpass human workers in efficiency~\citep{gilardi2023chatgpt}.
Building on this insight, we introduce \imagetotextrelevance, which measures the alignment between each textual step and its corresponding visual depiction, and \planscore, which evaluates whether a generated textual plan is both task-consistent and logically coherent.
In practice, \planscore completes evaluation in an average of $0.7$ seconds per task, compared to approximately $5$ minutes for human annotators. For complex tasks (\eg, \enquote{\textit{How to weave a rag rug?}}), human evaluation takes a longer time due to domain-specific knowledge requirements. Similarly, for \imagetotextrelevance, each step-level evaluation completes in about $1$ minute, representing a $66\%$ reduction in time compared to human assessment, which averages $3$ minutes. These results highlight the efficiency and practicality of LLM-based evaluation for multimodal planning tasks.

\noindent \textbf{\planscore.} Prior approaches to textual plan evaluation often rely on semantic similarity to reference plans~\citep{lu2023multimodal}, which may not fully capture planning accuracy or temporal coherence. We propose \planscore, a reference-free method that prompts a language model~\citep{brown2020language} to assess how well a generated plan aligns with the overall task goal. The prompt guides the model to consider both procedural correctness and logical step ordering. Empirically, we find that higher \planscore values correspond to more coherent plans that accurately reflect the intended procedure.

\noindent \textbf{\imagetotextrelevance.}
Cross-modal alignment between textual and visual plans is often evaluated using similarity-based metrics such as CLIPScore~\citep{hessel-etal-2021-clipscore} or Sentence-BERT~\citep{reimers2019sentence}, computed between generated visual captions and reference textual plans~\citep{lu2023multimodal}. While effective for coarse semantic matching, these methods may fail to capture finer-grained alignment, such as implicit object state changes not explicitly described in the text. To address this, we leverage multimodal language models, which have shown strong cross-modal reasoning capabilities~\citep{zhu2023minigpt, liu2023improvedllava}.
Inspired by recent work on text-to-image evaluation~\citep{huang2024t2i}, we employ a VLM with Chain-of-Thought (CoT) to assess cross-modal alignment at the step level. Specifically, we prompt the model to describe the contents of a generated image and compare the resulting description with the corresponding step text, evaluating alignment in terms of depicted actions and object states. We refer to the resulting score as \imagetotextrelevance.

\begin{table*}[t!]
\centering
\vspace{-0.1cm}
\resizebox{0.99\textwidth}{!}{%
\begin{tabular}{l cccccccc}
\toprule
\multirow{3}{*}{\textbf{Model}}& \multicolumn{4}{c}{\textbf{\textsc{RecipePlan}}} & \multicolumn{4}{c}{\textbf{\textsc{Wikiplan}}} \\
\cmidrule(lr){2-5}\cmidrule(lr){6-9}\
 & \planscore $\uparrow$& S-BERT $\uparrow$& WMD $\uparrow$& METEOR  
 $\uparrow$& \planscore $\uparrow$& S-BERT $\uparrow$& WMD $\uparrow$& METEOR $\uparrow$\\
    \midrule
    Text-Ref+SD & 73.85 & 0.72 & 0.13 & 0.09  & 62.53 & \textbf{0.78} & 0.70 & 0.20\\
    GPT-3.5+SD & 80.43 & 0.73 & 0.86 & 0.14 & 81.93 & 0.75 & \textbf{0.77} & 0.09\\ 
    TIP & 82.00 & 0.73 & 0.86 & 0.14 & 
    83.00 & \textbf{0.78} & \textbf{0.77} & 0.09 \\
    \rowcolor{lightgray!30}\textbf{\modelname} & \textbf{82.05} &\textbf{0.78}&\textbf{0.88}&\textbf{0.15}
    & \textbf{84.43} &0.77&0.75&\textbf{0.23}\\
    
    \bottomrule
\end{tabular}
}
\vspace{-0.3cm}
\caption{\textbf{Textual Evaluation on \textsc{RecipePlan} and \textsc{WikiPlan}.} Across both datasets, \modelnamenc consistently surpasses or achieves competitive performance against baselines.} 
\label{tab:textual_eval}
\end{table*}

\noindent \textbf{\visualsequenceordering.} 
To assess the informativeness and temporal coherence of visual plans, we additionally introduce a visual step reordering task. Given an unordered sequence of visual steps, the objective is to recover their correct temporal order. This task provides a direct measure of how well the visual outputs capture procedural structure. 
A related task was proposed by \citet{wu2022understanding}, who introduced multimodal instruction sequencing involving both textual and visual inputs. However, we find that including text in the sequence biases evaluation toward the textual modality, limiting sensitivity to visual quality (Appendix~\ref{sec:abl_metrics}). 
To address this, our formulation focuses solely on the visual modality, enabling the evaluation of visual procedural understanding independent of textual cues.
We adopt a pretrained visual sequencing model~\citep{wu2022understanding} that consists of a CLIP image encoder and an order decoder based on the BERSON framework~\citep{cui2018deep}. The vision encoder is trained with self-supervised objectives such as masked language modeling, patch-based image swapping, and sequential masked region modeling. For each visual plan, we randomly shuffle the step order and use the model to predict the correct sequence. Figure~\ref{fig:teaser} illustrates an example.

\section{Experiments}

We evaluate \modelnamenc on the \textsc{RecipePlan} and \textsc{WikiPlan}~\citep{lu2023multimodal} datasets. \textsc{RecipePlan} consists of 1,000 recipe tasks adapted from RecipeQA~\citep{yagcioglu2018recipeqa}, where each task includes a goal (taken from the recipe title) and a sequence of text-image pairs representing procedural steps. \textsc{WikiPlan} contains 1,000 tasks sourced from WikiHow articles, where the article title serves as the goal, the main body text forms the textual plan, and accompanying images comprise the visual plan.
We conduct experiments comparing \textbf{\modelname} with \textbf{TIP}~\citep{lu2023multimodal}, a dual prompting MPP method that integrates procedural knowledge from LLMs and T2I models by prompting both twice during inference. We also compare against two baselines from TIP: (1) \textbf{GPT-3.5+SD}, which independently generates textual plans using GPT-3.5 and visual plans using Stable Diffusion (SD); and (2) \textbf{Text-Ref+SD}, which generates images with Stable Diffusion (SD) from brief step titles instead of detailed steps.

Our evaluation is structured across three dimensions:
(i) \textbf{textual planning}, which assesses the accuracy and coherence of the generated textual plan;
(ii) \textbf{cross-modal alignment}, which evaluates the relevance between each visual step and its corresponding text; and
(iii) \textbf{visual ordering}, which measures the temporal consistency and informativeness of the visual plan. Implementation details can be found in Appendix~\ref{sup:implementation_detail}.

\subsection{Quantitative Evaluation}\label{subsec:quantitative_evaluation}

\noindent \textbf{Textual Evaluation.}
We employ \planscore to assess planning accuracy, alongside standard text similarity metrics {Sentence-BERT (S-BERT)}~\citep{reimers2019sentence}, {Word Movers Distance (WMD)}~\citep{kusner2015word}, and {METEOR}~\citep{banerjee-lavie-2005-meteor}. 
Table~\ref{tab:textual_eval} compares \modelnamenc against baselines. 
Overall, \modelnamenc achieves strong performance across both datasets. On \planscore, GPT-3.5+SD, TIP, and \modelnamenc perform similarly, reflecting the effectiveness of LLMs in producing coherent goal-aligned textual plans. In contrast, Text-Ref+SD performs worse due to the limited information available in step titles used as input.
Unlike reference-based metrics, \planscore does not rely on a fixed ground-truth sequence. Instead, it provides a reference-free assessment of how well the generated plan aligns with the task goal, accommodating multiple valid solution paths. On feature similarity metrics (S-BERT, WMD, METEOR), \modelnamenc consistently outperforms baselines, particularly on \textsc{RecipePlan}, indicating strong semantic alignment with the reference plans.
\begin{table}[t!]
\centering
\resizebox{\columnwidth}{!}{%
\begin{tabular}{l cccc}
\toprule
\multirow{3}{*}{\textbf{Model}}& \multicolumn{2}{c}{\textbf{\textsc{RecipePlan}}} & \multicolumn{2}{c}{\textbf{\textsc{Wikiplan}}} \\
\cmidrule(lr){2-3}\cmidrule(lr){4-5}\
 & \imagetotextrelevance $\uparrow$& CLIPScore   
 $\uparrow$& \imagetotextrelevance $\uparrow$ & CLIPScore $\uparrow$\\
    \midrule
    Text-Ref+SD & 70.81 & 60.68 & 61.61 & 65.42 \\
    GPT-3.5+SD & 71.49 & 73.00 & 63.18 & 71.08 \\
    TIP & 67.68 & 73.09 & 63.30 & 72.17 \\
    \rowcolor{lightgray!30}\textbf{\modelname} & \textbf{77.07} &\textbf{77.44}& \textbf{69.23} & \textbf{76.10}\\
    \bottomrule
\end{tabular}
}
\vspace{-0.3cm}
\caption{\textbf{Cross-Modal Step-level Evaluation on \textsc{RecipePlan} and \textsc{WikiPlan}.} \modelnamenc improves cross-modal alignment between visual and textual steps.}
\label{tab:cross_modal_evaluation}
\end{table}

\begin{table*}[t!]
\centering
\resizebox{0.99\linewidth}{!}{%
\begin{tabular}{lcccccccccccc}
\toprule
\multirow{2}{*}{\textbf{Model}} 
& \multicolumn{6}{c}{\textbf{\textsc{RecipePlan}}}
& \multicolumn{6}{c}{\textbf{\textsc{WikiPlan}}} \\
\cmidrule(lr){2-7} \cmidrule(lr){8-13}
& Acc $\uparrow$ & LCS $\uparrow$ & $\tau$ $\uparrow$ & Dist. $\downarrow$ & MS $\downarrow$ & WMS $\downarrow$
& Acc $\uparrow$ & LCS $\uparrow$ & $\tau$ $\uparrow$ & Dist. $\downarrow$ & MS $\downarrow$ & WMS $\downarrow$ \\
\midrule
Text-Ref+SD 
& 22.60 & 2.09 & 0.04 & 7.76 & 2.66 & 6.07
& 19.70 & 2.80 & 0.01 & 7.87 & 2.78 & 6.43 \\
GPT-3.5+SD 
& 21.65 & 1.83 & 0.03 & 7.78 & 2.63 & 5.99
& 18.95 & 2.69 & 0.01 & 7.66 & 2.67 & 6.23 \\
TIP 
& 21.70 & 1.81 & 0.05 & 7.79 & 2.66 & 5.99
& 18.79 & 2.83 & 0.02 & 7.92 & 2.75 & 6.30 \\
\rowcolor{lightgray!30}
\textbf{\modelname} 
& \textbf{27.50} & \textbf{3.09} & \textbf{0.22} & \textbf{6.51} & \textbf{2.39} & \textbf{4.99}
& \textbf{23.43} & \textbf{2.91} & \textbf{0.05} & \textbf{7.60} & \textbf{2.60} & \textbf{5.90} \\
\bottomrule
\end{tabular}
}
\vspace{-0.3cm}
\caption{\textbf{Visual Sequence Ordering (\visualsequenceordering) Evaluation.} MMPlanner consistently outperforms baselines.}
\label{tab:visual_sequence_ordering}
\end{table*}

\noindent \textbf{Cross-modal Step-Level Evaluation.} We evaluate step-level cross-modal alignment using {CLIPScore}~\citep{hessel-etal-2021-clipscore} and our proposed {\imagetotextrelevance}. 
As shown in Table~\ref{tab:cross_modal_evaluation}, \modelnamenc outperforms all baselines on both datasets. On \imagetotextrelevance, \modelnamenc improves over TIP by $11.9\%$ on \textsc{RecipePlan} and $9.37\%$ on \textsc{WikiPlan}. 
Additionally, our Cross-modal Step Image Selector yields CLIPScore improvements of $6.1\%$ and $5.5\%$ over TIP on \textsc{RecipePlan} and \textsc{WikiPlan}, respectively. These results demonstrate that \modelnamenc produces visual plans that are more semantically aligned with their corresponding textual steps. 
Importantly, \imagetotextrelevance shows stronger correlation with human ratings ($\rho=0.57$) compared to CLIPScore ($\rho=0.37$) (details in Appendix~\ref{subsec:ca_score_correlation}), underscoring the reliability of our proposed metric.

\noindent \textbf{Visual Ordering.} 
We evaluate the temporal coherence of generated visual step sequences on the \visualsequenceordering task with six established ordering metrics: {Accuracy (Acc)}, {Distance (Dist)}, {Minimum Swap (MS)}, {Weighted Minimum Swap (WMS)}, {Longest Common Subsequence (LCS)}, and {Kendall’s Tau ($\bm{\tau}$)}~\citep{lapata2003probabilistic}. Detailed metric definitions can be found in Appendix~\ref{sec:ev_metrics}.
Table~\ref{tab:visual_sequence_ordering} reports results on both datasets. On \textsc{RecipePlan}, \modelnamenc outperforms all methods by substantial margins, achieving gains of $26.7\%$, $16.4\%$, $10.15\%$, and $16.7\%$ over the second-best method (TIP) on Accuracy, Dist, MS, and WMS, respectively. 
On \textsc{WikiPlan}, \modelnamenc shows consistent improvements over TIP with relative gains of $24.7\%$, $4.0\%$, $5.5\%$, and $6.3\%$ on the same metrics. On \textsc{RecipePlan}, \modelnamenc surpasses TIP by $47.85\%$ in LCS and by over $340\%$ in Kendall’s Tau, further indicating stronger global temporal consistency in the generated visual plans.

\noindent \textbf{Inference Comparison.} 
TIP requires two sequential prompts for text-to-image and image-to-text models, resulting in increased inference time. In contrast, \modelnamenc integrates reasoning over previous steps and object states directly via OSR-CoT, eliminating the need for dual prompting and significantly streamlining inference. As a result, \modelnamenc achieves an average inference time of $52.02$ seconds, compared to $96.77$ seconds for TIP — a relative reduction of approximately $46.25\%$. 

\subsection{Ablation Studies}\label{subsec:ablation_studies}
We conduct ablation studies to analyze the contributions of \modelnamenc components (Section~\ref{subsec:ablation_studies} and Appendix~\ref{sup:ablplanner}). We evaluate:
(1) the impact of each \modelnamenc module;
(2) the importance of different components within the OSR-CoT prompt;
(3) the effectiveness of BLIP-2 as a cross-modal feature extractor;
(4) the influence of the sampling hyperparameter $K$ in visual selection; and
(5) the role of the visual goal in textual plan generation.
We further validate our proposed evaluation by:
(1) assessing the reliability of \planscore;
(2) evaluating the correlation between \imagetotextrelevance and human judgments; and
(3) analyzing robustness across different LLMs/VLMs (Appendix~\ref{sec:abl_metrics}).

\begin{figure*}[t!]
\centering
  \includegraphics[width=0.99\textwidth]{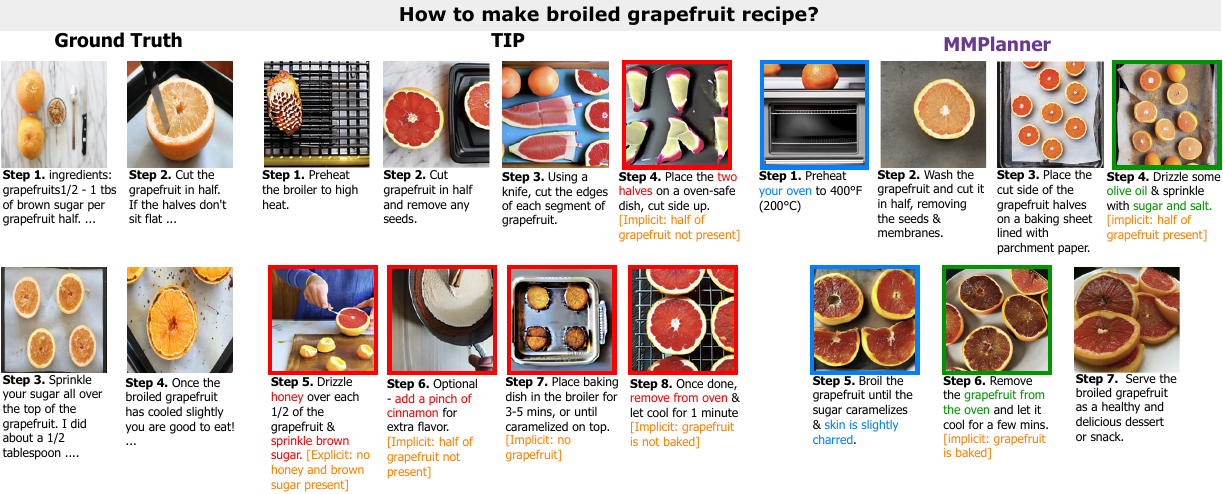}
  \vspace{-0.3cm}
\caption{\textbf{Qualitative Comparison TIP vs. \modelnamenc on \textsc{RecipePlan} task \enquote{How to make broiled grapefruit recipe}.}
Explicit state changes are those clearly described in the textual step and expected in the visual step. Implicit state changes are not explicitly stated in the step text but are necessary to convey in the visual step.
\underline{Left:} TIP fails to accurately reflect both explicit and implicit object state changes in visual steps (\textcolor{red}{red text and bboxes}).
\underline{Right:} \modelnamenc captures both explicit (\textcolor{blue}{blue}) and implicit (\textcolor{green}{green}) object state changes. Explanations in \textcolor{orange}{orange} text.}
  \label{fig:qualitative_recipe_1}
\end{figure*}
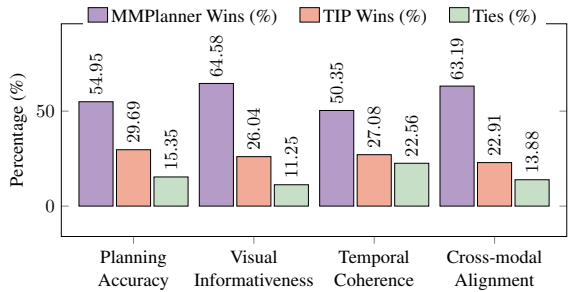
\begin{figure}[t!]
    \centering
    \resizebox{0.97\columnwidth}{!}{%
    \begin{tikzpicture}
        \begin{axis}[
            ybar,
            ymin=0, ymax=80,
            symbolic x coords={
                Planning\\Accuracy, 
                Visual\\Informativeness, 
                Temporal\\Coherence, 
                Cross-modal\\Alignment
            },
            xtick=data,
            xticklabel style={align=center},
            nodes near coords,
            nodes near coords style={rotate=90, anchor=west},
            enlargelimits=0.2,
            bar width=20pt,
            width=12cm, height=6cm,
            legend style={at={(0.5,1.1)}, anchor=north, legend columns=-1, draw=none},
            legend image code/.code={\draw[#1,draw=black] (0cm,-0.1cm) rectangle (0.4cm,0.2cm);},
            ylabel={Percentage (\%)},
            ylabel style={align=center, yshift=-10pt},
        ]
            \addplot[fill=mmPurple!50] coordinates {(Planning\\Accuracy,54.95) (Visual\\Informativeness,64.58) (Temporal\\Coherence,50.35) (Cross-modal\\Alignment,63.19)};
            \addplot[fill=burntsienna!60] coordinates {(Planning\\Accuracy,29.69) (Visual\\Informativeness,26.04) (Temporal\\Coherence,27.08) (Cross-modal\\Alignment,22.91)};
            \addplot[fill=darkseagreen!50] coordinates {(Planning\\Accuracy,15.35) (Visual\\Informativeness,11.25) (Temporal\\Coherence,22.56) (Cross-modal\\Alignment,13.88)};
            \legend{MMPlanner Wins (\%)~~~~~, TIP Wins (\%)~~~~~, Ties (\%)}
        \end{axis}
    \end{tikzpicture}
    }
    \vspace{-0.3cm}
    \caption{\textbf{User Study \modelnamenc vs. TIP.} Multimodal plans evaluated by participants across four dimensions. Bars present percentage of wins and ties.}
    \label{fig:human_evalaution}
    \vspace{-0.3cm}
\end{figure}
\begin{table}[t!]
\centering
\resizebox{\linewidth}{!}{%
\begin{tabular}{l cccccc}
\toprule
\textbf{Model} & \imagetotextrelevance $\uparrow$ & CLIPScore $\uparrow$ & Acc $\uparrow$ & Dist.$\downarrow$ & MS $\downarrow$ & WMS $\downarrow$ \\
\midrule

\rowcolor{green!8}
\multicolumn{7}{c}{\textbf{\textsc{RecipePlan}}} \\
LLaVa+SD  & 72.02 & 73.87 & 22.10 & 7.71 & 2.70 & 6.41 \\
\quad + OSR-CoT  & 75.15 & 75.12 & 25.50 & 7.03 & 2.53 & 5.16 \\
\quad + Previous Steps  & 75.27 & 75.19 & 26.80 & 6.82 & 2.48 & 5.03 \\
\cellcolor{lightgray!30}\quad + CM Sel. (\modelname) & \cellcolor{lightgray!30}\textbf{77.07} & \cellcolor{lightgray!30}\textbf{77.55} & \cellcolor{lightgray!30}\textbf{27.50} & \cellcolor{lightgray!30}\textbf{6.51} & \cellcolor{lightgray!30}\textbf{2.39} & \cellcolor{lightgray!30}\textbf{4.99} \\
\cmidrule(lr){1-7}

\rowcolor{green!8}
\multicolumn{7}{c}{\textbf{\textsc{WikiPlan}}} \\
LLaVa+SD  & 71.65 & 72.83 & 22.01 & 7.70 & 2.62 & 6.06 \\
\quad + OSR-CoT & 74.58 & 74.27 & 22.29 & 7.66 & 2.61 & 5.98 \\
\quad + Previous Steps  & 75.70 & 74.35 & 23.17 & 7.78 & 2.60 & 6.16 \\
\cellcolor{lightgray!30}\quad + CM Sel. (\modelname) & \cellcolor{lightgray!30}\textbf{77.48} & \cellcolor{lightgray!30}\textbf{76.10} & \cellcolor{lightgray!30}\textbf{23.43} & \cellcolor{lightgray!30}\textbf{7.60} & \cellcolor{lightgray!30}\textbf{2.60} & \cellcolor{lightgray!30}\textbf{5.90} \\

\bottomrule
\end{tabular}
}
\vspace{-0.3cm}
\caption{\textbf{Ablation on \modelnamenc Components.} \modelnamenc's components collectively improve cross-modal alignment and temporal coherence.}
\label{tab:ablation_studies_components}
\end{table}

\noindent \textbf{\modelnamenc Components.}
We conduct an ablation to evaluate the contribution of each module in \modelnamenc. We begin with a baseline variant, \textbf{LLaVa+SD}, where textual plans are generated using LLaVa and directly passed to Stable Diffusion (SD) to produce visual steps. We then integrate the Image Description Generator component with Object State Reasoning via Chain-of-Thought prompting (\textbf{OSR-CoT}), without conditioning on previous steps. Next, we incorporate prior step context into OSR-CoT (\textbf{Previous Steps}), followed by the integration of our Cross-modal Step Image Selection module (\textbf{CM Sel.}).
As shown in Table~\ref{tab:ablation_studies_components}, each component incrementally improves performance, with the full model (\modelnamenc) achieving the highest scores in both cross-modal alignment and \visualsequenceordering. Notably, the introduction of OSR-CoT leads to a substantial gain in \imagetotextrelevance on both \textsc{RecipePlan} and \textsc{WikiPlan}, underscoring the value of structured image descriptions. While improvements on \visualsequenceordering are more modest for \textsc{WikiPlan}, the inclusion of OSR-CoT, previous step context, and CM Sel. still results in consistent gains across Acc., Dist., and WMS.

\subsection{Human Evaluation}\label{sec:human_evaluation}
We conduct a user study to compare \modelnamenc and TIP across four key dimensions:
\textbf{1) Planning Accuracy}, \ie, whether following the multimodal plan would successfully complete the task;
\textbf{2) Visual Informativeness}, \ie, how well the visual steps support task execution;
\textbf{3) Temporal Coherence}, \ie, whether the steps are presented in a logical order; and
\textbf{4) Cross-modal Alignment}, \ie, the consistency between each step image and its corresponding textual step.
We conducted a human evaluation with 26 participants who assessed multimodal plans generated by TIP and \modelnamenc across 12 distinct tasks. Each participant compared plans from both models across four key dimensions, resulting in a total of 1,248 pairwise judgments. For each task, participants were presented with two unlabeled step-by-step multimodal plans, one from each model, alongside the high-level task objective and were asked to choose their preferred plan based on four criteria: (i) accuracy of the steps in achieving the task goal; (ii) visual informativeness of each step; (iii) temporal coherence across steps; and (iv) alignment between textual and visual modalities. This setup ensures an unbiased and comprehensive evaluation of plan quality.
As shown in Figure~\ref{fig:human_evalaution}, \modelnamenc receives consistently higher preference across all evaluation criteria, demonstrating improved planning accuracy, visual clarity, temporal structure, and visual-text consistency compared to TIP.

\subsection{Qualitative Evaluation}\label{subsec:qualitative_evaluation}
We compare multimodal plans generated by TIP and \modelnamenc. Figure~\ref{fig:qualitative_recipe_1} presents example step-by-step plans for the task \enquote{\textit{How to make broiled grapefruit recipe}}. TIP's visual steps often fail to capture key object state changes, both explicit and implicit, as described in the textual instructions. This observation aligns with its lower \imagetotextrelevance scores. In contrast, \modelnamenc generates visual steps that more accurately reflect the described explicit (highlighted with blue color) and implicit (highlighted with green color) object states. Furthermore, the visual plan produced by \modelnamenc is qualitatively closer to the ground truth plan, demonstrating better alignment between text and image modalities and stronger procedural understanding. Additional examples in Appendix \ref{sec:qualexamples}.

\section{Broader Impacts}
This work aims to advance the deployment of multimodal generative models, such as LLMs and text-to-image models, for real-world, step-by-step task assistance. Our goal is to make task-driven assistive technology more practical and accessible, particularly for users who benefit from visual guidance.
We acknowledge the limitations of generative models, especially their susceptibility to hallucination and misinformation. To address this, OSR-CoT encourages grounded reasoning by decomposing tasks into smaller, verifiable steps, reducing the risk of unsupported outputs. Future directions could focus on integrating external knowledge verification to further enhance the reliability and trustworthiness of AI-generated multimodal plans.

\section{Limitations}
\modelnamenc leverages LLMs and VLMs for multimodal plan generation and evaluation via \planscore and \imagetotextrelevance. However, hallucinations remain a known limitation of LLMs~\citep{xu2024hallucination}. This issue is most evident before applying OSR-CoT (Section~\ref{subsec:image_description_generator}). For example, the generated description of the task \enquote{\textit{How to make cheese garlic pull-apart bread?}} (Figure~\ref{fig:qualitative_recipe_2}) for step 5
is \enquote{\textit{A bowl of dough mixture is forming. Flour and butter can be seen in the background}} without OSR-CoT, showing that the LLM introduces unrelated details about flour and butter from earlier steps. In contrast, OSR-CoT yields \enquote{\textit{The milk mixture being slowly stirred into the dry ingredients}}, more accurately aligning with the step's intent and focusing on what is relevant (ingredients, processes, \etc) for that specific step.
This shows how OSR-CoT reduces hallucinations and improves step relevance. 
OSR-CoT improves object state reasoning, but \modelnamenc does not explicitly enforce visual consistency for peripheral elements such as cookingware shape or material. For instance, in Figure~\ref{fig:qualitative_recipe_2}, the bowl depicted in steps 2, 3, and 5 varies in appearance.
\modelnamenc inherits limitations from Stable Diffusion, particularly its inability to render concepts absent from its training data. For example, given the step \enquote{\textit{Beat the egg with a fork}} from the task \enquote{\textit{How to make an omelet}}, the model fails to generate an accurate depiction of a \enquote{beaten egg}.  Addressing such inconsistencies remains an avenue for future work.
Finally, while \planscore shows consistent monotonic trends under plan degradations, its absolute calibration can be imperfect. In some cases, low-quality plans (\eg with many deleted steps) still receive high scores. This reflects a broader limitation of LLM-based evaluators, where prompt adherence and human alignment are not guaranteed. Future work could improve calibration through human feedback, refined prompts, or preference-based fine-tuning. As multimodal LLMs continue to improve, they offer potential for better MPP evaluation frameworks, but further work is needed to refine both generation and evaluation.

\section{Conclusion}
We present \modelnamenc, a zero-shot multimodal procedural planning method using OSR-CoT prompting to capture explicit and implicit object state changes. To evaluate generated plans, we propose an automatic evaluation that assesses planning accuracy, cross-modal alignment, and temporal coherence. 
Experiments show \modelnamenc generates accurate and coherent multimodal plans.

\section{Acknowledgments}
This research is based on work partially supported by the Amazon–Virginia Tech Initiative for Efficient and Robust Machine Learning and by the U.S. Defense Advanced Research Projects Agency (DARPA) under award numbers HR00112390062 and HR001125C0303. The views and conclusions contained herein are those of the authors and should not be interpreted as necessarily representing the official policies, either expressed or implied, of Amazon, DARPA, or the U.S. Government. The U.S. Government is authorized to reproduce and distribute reprints for governmental purposes notwithstanding any copyright annotation therein.

{\small
\bibliography{biblio}

\begin{thebibliography}{39}
\providecommand{\natexlab}[1]{#1}

\bibitem[{Alikhani et~al.(2019)Alikhani, Nag~Chowdhury, de~Melo, and Stone}]{alikhani-etal-2019-cite}
Malihe Alikhani, Sreyasi Nag~Chowdhury, Gerard de~Melo, and Matthew Stone. 2019.
\newblock {CITE}: A corpus of image-text discourse relations.
\newblock In \emph{Annual Conference of the North American Chapter of the Association for Computational Linguistics (ACL).}

\bibitem[{Ashutosh et~al.(2023)Ashutosh, Ramakrishnan, Afouras, and Grauman}]{ashutosh2023video}
Kumar Ashutosh, Santhosh~Kumar Ramakrishnan, Triantafyllos Afouras, and Kristen Grauman. 2023.
\newblock Video-mined task graphs for keystep recognition in instructional videos.
\newblock In \emph{Advances in Neural Information Processing Systems (NeurIPS)}.

\bibitem[{Banerjee and Lavie(2005)}]{banerjee-lavie-2005-meteor}
Satanjeev Banerjee and Alon Lavie. 2005.
\newblock {METEOR}: An automatic metric for {MT} evaluation with improved correlation with human judgments.
\newblock In \emph{{ACL} Workshop on Intrinsic and Extrinsic Evaluation Measures for Machine Translation and/or Summarization}.

\bibitem[{Brown et~al.(2020)Brown, Mann, Ryder, Subbiah, Kaplan, Dhariwal, Neelakantan, Shyam, Sastry, Askell et~al.}]{brown2020language}
Tom Brown, Benjamin Mann, Nick Ryder, Melanie Subbiah, Jared~D Kaplan, Prafulla Dhariwal, Arvind Neelakantan, Pranav Shyam, Girish Sastry, Amanda Askell, et~al. 2020.
\newblock Language models are few-shot learners.
\newblock In \emph{Advances in Neural Information Processing Systems (NeurIPS)}.

\bibitem[{Chen et~al.(2017)Chen, Liu, Yin, and Tang}]{chen2017survey}
Hongshen Chen, Xiaorui Liu, Dawei Yin, and Jiliang Tang. 2017.
\newblock A survey on dialogue systems: Recent advances and new frontiers.
\newblock \emph{{ACM SIGKDD Explorations Newsletter}}.

\bibitem[{Cui et~al.(2018)Cui, Li, Chen, and Zhang}]{cui2018deep}
Baiyun Cui, Yingming Li, Ming Chen, and Zhongfei Zhang. 2018.
\newblock Deep attentive sentence ordering network.
\newblock In \emph{Conference on Empirical Methods in Natural Language Processing (EMNLP)}.

\bibitem[{Gilardi et~al.(2023)Gilardi, Alizadeh, and Kubli}]{gilardi2023chatgpt}
F.~Gilardi, M.~Alizadeh, and M.~Kubli. 2023.
\newblock Chatgpt outperforms crowd workers for text-annotation tasks.
\newblock \emph{National Academy of Sciences}.

\bibitem[{Hessel et~al.(2021)Hessel, Holtzman, Forbes, Le~Bras, and Choi}]{hessel-etal-2021-clipscore}
Jack Hessel, Ari Holtzman, Maxwell Forbes, Ronan Le~Bras, and Yejin Choi. 2021.
\newblock {CLIPS}core: A reference-free evaluation metric for image captioning.
\newblock In \emph{Conference on Empirical Methods in Natural Language Processing (EMNLP)}.

\bibitem[{Huang et~al.(2024)}]{huang2024t2i}
K.~Huang et~al. 2024.
\newblock {T2I}-compbench: A comprehensive benchmark for open-world compositional text-to-image generation.
\newblock In \emph{Advances in Neural Information Processing Systems (NeurIPS)}.

\bibitem[{Inan et~al.(2021)Inan, Sharma, Khalid, Soricut, Stone, and Alikhani}]{inan-etal-2021-cosmic}
Mert Inan, Piyush Sharma, Baber Khalid, Radu Soricut, Matthew Stone, and Malihe Alikhani. 2021.
\newblock Cosmic: A coherence-aware generation metric for image descriptions.
\newblock In \emph{Findings of the Association for Computational Linguistics (EMNLP)}.

\bibitem[{Kovalchuk et~al.(2021)Kovalchuk, Shekhar, and Brafman}]{kovalchuk2021verifying}
Alexander Kovalchuk, Shashank Shekhar, and Ronen~I Brafman. 2021.
\newblock Verifying plans and scripts for robotics tasks using performance level profiles.
\newblock In \emph{International Conference on Automated Planning and Scheduling}.

\bibitem[{Kusner et~al.(2015)Kusner, Sun, Kolkin, and Weinberger}]{kusner2015word}
Matt Kusner, Yu~Sun, Nicholas Kolkin, and Kilian Weinberger. 2015.
\newblock From word embeddings to document distances.
\newblock In \emph{International Conference on Machine Learning (ICML)}.

\bibitem[{Lapata(2003)}]{lapata2003probabilistic}
Mirella Lapata. 2003.
\newblock Probabilistic text structuring: Experiments with sentence ordering.
\newblock In \emph{Association for Computational Linguistics (ACL)}.

\bibitem[{Li et~al.(2023)}]{li2023blip}
J.~Li et~al. 2023.
\newblock Blip-2: Bootstrapping language-image pre-training with frozen image encoders and large language models.
\newblock In \emph{International Conference on Machine Learning (ICML)}.

\bibitem[{Liu et~al.(2023)Liu, Li, Li, and Lee}]{liu2023improvedllava}
Haotian Liu, Chunyuan Li, Yuheng Li, and Yong~Jae Lee. 2023.
\newblock Improved baselines with visual instruction tuning.
\newblock In \emph{NeurIPS Workshop on Instruction Tuning and Instruction Following}.

\bibitem[{Lu et~al.(2022)Lu, Feng, Zhu, Xu, Wang, Eckstein, and Wang}]{lu2022neuro}
Yujie Lu, Weixi Feng, Wanrong Zhu, Wenda Xu, Xin~Eric Wang, Miguel Eckstein, and William~Yang Wang. 2022.
\newblock Neuro-symbolic procedural planning with commonsense prompting.
\newblock In \emph{International Conference on Learning Representations (ICLR)}.

\bibitem[{Lu et~al.(2024)Lu, Lu, Chen, Zhu, Wang, and Wang}]{lu2023multimodal}
Yujie Lu, Pan Lu, Zhiyu Chen, Wanrong Zhu, Xin Wang, and William~Yang Wang. 2024.
\newblock Multimodal procedural planning via dual text-image prompting.
\newblock In \emph{Findings of the Association for Computational Linguistics (EMNLP)}.

\bibitem[{Lyu et~al.(2021)Lyu, Zhang, and Chris}]{lyu2021goal}
Qing Lyu, Li~Zhang, and Callison-Burch Chris. 2021.
\newblock Goal-oriented script construction.
\newblock In \emph{International Conference on Natural Language Generation}.

\bibitem[{Menon and Vondrick(2022)}]{menon2022visual}
Sachit Menon and Carl Vondrick. 2022.
\newblock Visual classification via description from large language models.
\newblock In \emph{International Conference on Learning Representations (ICLR)}.

\bibitem[{Niu et~al.(2024)Niu, Guo, Chen, Lin, and Chang}]{niu2024schema}
Yulei Niu, Wenliang Guo, Long Chen, Xudong Lin, and Shih-Fu Chang. 2024.
\newblock Schema: State changes matter for procedure planning in instructional videos.
\newblock In \emph{International Conference on learning Representations (ICLR)}.

\bibitem[{Ouyang et~al.(2022)Ouyang, Wu, Jiang, Almeida, Wainwright, Mishkin, Zhang, Agarwal, Slama, Ray et~al.}]{ouyang2022training}
Long Ouyang, Jeffrey Wu, Xu~Jiang, Diogo Almeida, Carroll Wainwright, Pamela Mishkin, Chong Zhang, Sandhini Agarwal, Katarina Slama, Alex Ray, et~al. 2022.
\newblock Training language models to follow instructions with human feedback.
\newblock In \emph{Advances in Neural Information Processing Systems (NeurIPS)}.

\bibitem[{Reimers and Gurevych(2019)}]{reimers2019sentence}
Nils Reimers and Iryna Gurevych. 2019.
\newblock Sentence-bert: Sentence embeddings using siamese bert-networks.
\newblock In \emph{Conference on Empirical Methods in Natural Language Processing and International Joint Conference on Natural Language Processing (EMNLP-IJCNLP)}.

\bibitem[{Rombach et~al.(2022)Rombach, Blattmann, Lorenz, Esser, and Ommer}]{rombach2022high}
Robin Rombach, Andreas Blattmann, Dominik Lorenz, Patrick Esser, and Bj\"orn Ommer. 2022.
\newblock High-resolution image synthesis with latent diffusion models.
\newblock In \emph{IEEE/CVF Conference on Computer Vision and Pattern Recognition (CVPR)}.

\bibitem[{Sedgwick(2014)}]{sedgwick2014spearman}
Philip Sedgwick. 2014.
\newblock Spearman's rank correlation coefficient.
\newblock \emph{BMJ: British Medical Journal}, 349.

\bibitem[{Song et~al.(2023)Song, Wu, Washington, Sadler, Chao, and Su}]{song2023llm}
Chan~Hee Song, Jiaman Wu, Clayton Washington, Brian~M Sadler, Wei-Lun Chao, and Yu~Su. 2023.
\newblock Llm-planner: Few-shot grounded planning for embodied agents with large language models.
\newblock In \emph{IEEE/CVF International Conference on Computer Vision (ICCV)}.

\bibitem[{Sou{\v{c}}ek et~al.(2025)Sou{\v{c}}ek, Gatti, Wray, Laptev, Damen, and Sivic}]{souvcek2024showhowto}
Tom{\'a}{\v{s}} Sou{\v{c}}ek, Prajwal Gatti, Michael Wray, Ivan Laptev, Dima Damen, and Josef Sivic. 2025.
\newblock Showhowto: Generating scene-conditioned step-by-step visual instructions.
\newblock In \emph{IEEE/CVF Conference on Computer Vision and Pattern Recognition (CVPR)}.

\bibitem[{Sun et~al.(2023)Sun, Xu, Zhai, and Ji}]{sun-etal-2023-incorporating}
Chenkai Sun, Tie Xu, Cheng~Xiang Zhai, and Heng Ji. 2023.
\newblock Incorporating task-specific concept knowledge into script learning.
\newblock In \emph{Conference of the European Chapter of the Association for Computational Linguistics (EACL)}.

\bibitem[{Taori et~al.(2023)Taori, Gulrajani, Zhang, Dubois, Li, Guestrin, Liang, and Hashimoto}]{taori2023stanford}
Rohan Taori, Ishaan Gulrajani, Tianyi Zhang, Yann Dubois, Xuechen Li, Carlos Guestrin, Percy Liang, and Tatsunori~B Hashimoto. 2023.
\newblock Stanford alpaca: An instruction-following llama model.

\bibitem[{Wang et~al.(2023)Wang, Li, Chan, Huang, Hockenmaier, Chowdhary, and Ji}]{wang-etal-2023-multimedia}
Qingyun Wang, Manling Li, Hou~Pong Chan, Lifu Huang, Julia Hockenmaier, Girish Chowdhary, and Heng Ji. 2023.
\newblock Multimedia generative script learning for task planning.
\newblock In \emph{Association for Computational Linguistics (ACL)}.

\bibitem[{Wei et~al.(2022)Wei, Wang, Schuurmans, Bosma, Xia, Chi, Le, Zhou et~al.}]{wei2022chain}
Jason Wei, Xuezhi Wang, Dale Schuurmans, Maarten Bosma, Fei Xia, Ed~Chi, Quoc~V Le, Denny Zhou, et~al. 2022.
\newblock Chain-of-thought prompting elicits reasoning in large language models.
\newblock In \emph{Advances in Neural Information Processing Systems (NeurIPS)}.

\bibitem[{Wu et~al.(2022)Wu, Spangher, Alipoormolabashi, Freedman, Weischedel, and Peng}]{wu2022understanding}
Te-Lin Wu, Alex Spangher, Pegah Alipoormolabashi, Marjorie Freedman, Ralph Weischedel, and Nanyun Peng. 2022.
\newblock Understanding multimodal procedural knowledge by sequencing multimodal instructional manuals.
\newblock In \emph{Association for Computational Linguistics (ACL)}.

\bibitem[{Xu et~al.(2024)Xu, Jain, and Kankanhalli}]{xu2024hallucination}
Ziwei Xu, Sanjay Jain, and Mohan Kankanhalli. 2024.
\newblock Hallucination is inevitable: An innate limitation of large language models.
\newblock \emph{arXiv:2401.11817}.

\bibitem[{Yagcioglu et~al.(2018)Yagcioglu, Erdem, Erdem, and Ikizler-Cinbis}]{yagcioglu2018recipeqa}
Semih Yagcioglu, Aykut Erdem, Erkut Erdem, and Nazli Ikizler-Cinbis. 2018.
\newblock Recipeqa: A challenge dataset for multimodal comprehension of cooking recipes.
\newblock In \emph{Conference on Empirical Methods in Natural Language Processing (EMNLP)}.

\bibitem[{Yoneda et~al.(2024)Yoneda, Fang, Li, Zhang, Jiang, Lin, Picker, Yunis, Mei, and Walter}]{yoneda2023statler}
Takuma Yoneda, Jiading Fang, Peng Li, Huanyu Zhang, Tianchong Jiang, Shengjie Lin, Ben Picker, David Yunis, Hongyuan Mei, and Matthew~R. Walter. 2024.
\newblock Statler: State-maintaining language models for embodied reasoning.
\newblock In \emph{Proceedings of the IEEE International Conference on Robotics and Automation (ICRA)}.

\bibitem[{Yuan et~al.(2023)Yuan, Chen, Fu, Ge, Shah, Jankowski, Xiao, and Yang}]{yuan-etal-2023-distilling}
Siyu Yuan, Jiangjie Chen, Ziquan Fu, Xuyang Ge, Soham Shah, Charles Jankowski, Yanghua Xiao, and Deqing Yang. 2023.
\newblock Distilling script knowledge from large language models for constrained language planning.
\newblock In \emph{Association for Computational Linguistics (ACL)}.

\bibitem[{Zhao et~al.(2023)Zhao, Lee, and Hsu}]{zhao2023large}
Zirui Zhao, Wee~Sun Lee, and David Hsu. 2023.
\newblock Large language models as commonsense knowledge for large-scale task planning.
\newblock In \emph{RSS Workshop on Learning for Task and Motion Planning}.

\bibitem[{Zhou et~al.(2022)}]{zhou2022show}
Shuyan Zhou et~al. 2022.
\newblock Show me more details: Discovering hierarchies of procedures from semi-structured web data.
\newblock In \emph{Association for Computational Linguistics (ACL)}.

\bibitem[{Zhou et~al.(2023)Zhou, Li et~al.}]{zhou-etal-2023-non}
Yu~Zhou, Sha Li, et~al. 2023.
\newblock Non-sequential graph script induction via multimedia grounding.
\newblock In \emph{Association for Computational Linguistics (ACL)}.

\bibitem[{Zhu et~al.(2023)Zhu, Chen, Shen, Li, and Elhoseiny}]{zhu2023minigpt}
Deyao Zhu, Jun Chen, Xiaoqian Shen, Xiang Li, and Mohamed Elhoseiny. 2023.
\newblock Minigpt-4: Enhancing vision-language understanding with advanced large language models.
\newblock In \emph{International Conference on Learning Representations (ICLR)}.

\end{thebibliography}
}
\clearpage
\newpage
\appendix
 \section{Implementation Details}\label{sup:implementation_detail}
We employ LLaVa-1.5-7B~\citep{liu2023improvedllava} and GPT-3.5~\citep{brown2020language} for the Textual Plan Generator (Section~\ref{subsec:textual_plan_generator}) and Image Description Generator (Section~\ref{subsec:image_description_generator}), respectively. For \planscore and \imagetotextrelevance, we utilize GPT-3.5~\citep{brown2020language} and MiniGPT-4~\citep{zhu2023minigpt}, respectively. 
Prompts are described below.

\noindent\textbf{Textual Plan Generator:}
Given a multimodal goal, we construct a prompt that asks the VLM to generate a textual step-by-step plan, \ie,
\begin{figure}[h!]
\vspace{-0.3cm}
\begin{tcolorbox}[colback=cverbbg, boxrule=-0.2pt, sharp corners, boxsep=0pt]
\begin{minipage}{1\linewidth}
\centering
  \includegraphics[width=0.6\linewidth]{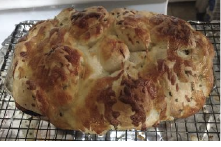}
  \end{minipage}
  \end{tcolorbox}
  \vspace{-0.5cm}
\end{figure}

\vspace{-1.1cm}
\begin{coloredlcverbatim}
Using the image as a reference, and goal  
"How to make garlic pull-apart bread?",   
give step-by-step brief instructions,  
according to the following format:
1. Start each step with the step number.
2. 1 sentence of 50 words maximum for each step.
\end{coloredlcverbatim}

\noindent\textbf{Image Description Generator.}
We introduce Chain-of-Thought with Object State Reasoning (OSR-CoT) prompting designed to generate descriptions for visual steps based on textual instructions. To manage token limitations in GPT-3.5, we cap the number of prior steps  ([prev\_steps]) used as background information to 10, closely matching the average number of ground truth steps (8.92). The prompt also includes an in-context example to guide the model's reasoning.
\noindent For example, the \textbf{OSR-CoT prompt} and 
\textcolor{maroon}{\textbf{in-context example}} 
for the \textcolor{purple}{\textbf{[goal]}} \enquote{How to make cheesy garlic pull-apart bread} and example \textcolor{orange}{\textbf{[step]}} \enquote{3. Add 1 cup of unsalted butter, cut into small pieces, and mix until the mixture is crumbly} is as follows:
\vspace{-0.3cm}
\begin{coloredlcverbatim}
First, describe details of \textcolor{orange}{\textbf{[step]}} for \textcolor{purple}{\textbf{[goal]}}
with one verb. Use \textcolor{teal}{\textbf{[prev_steps]}} for 
background information.
Second, use 3 sentences to describe the state
changes of objects before and after \textcolor{orange}{\textbf{[step]}}, 
avoiding using \textcolor{violet}{\textbf{[verb]}}.
Third, write description of the \textcolor{orange}{\textbf{[step]}} image
containing the 3 sentences after \textcolor{orange}{\textbf{[step]}}.
\textcolor{maroon}{\textbf{[goal]:} Task: How to make fried egg with cheese.}
\textcolor{maroon}{\textbf{[step]:} 3. Pour a small amount of butter} 
\textcolor{maroon}{        or oil into the pan.}
\textcolor{maroon}{\textbf{[prev_steps]:} 1. Crack an egg into a bowl.} 
\textcolor{maroon}{              2. Heat a non-stick frying pan} 
\textcolor{maroon}{              on medium heat.}

\textcolor{maroon}{\textbf{Description:}} 
\textcolor{maroon}{Pour a small amount of butter or oil into a pan.} 
\textcolor{maroon}{\textbf{Before:}}
\textcolor{maroon}{- An egg is cracked in a bowl.}
\end{coloredlcverbatim}
\begin{coloredlcverbatim}
\textcolor{maroon}{- Non-stick frying pan is heated on medium heat.}
\textcolor{maroon}{- The pan is empty without any butter/oil in it.}
\textcolor{maroon}{\textbf{After:}}
\textcolor{maroon}{-The butter or oil is in the pan.} 
\textcolor{maroon}{- The pan is coated with butter or oil.}
\textcolor{maroon}{- The pan is ready for cooking the egg.} 
\textcolor{maroon}{\textbf{Image Description:}}
\textcolor{maroon}{A non-stick frying pan with butter or oil}
\textcolor{maroon}{poured into it.} 

\textcolor{purple}{\textbf{[goal]}}: How to make a cheesy garlic pull-apart 
        bread?
\textcolor{orange}{\textbf{[step]}}: 3. Add 1 cup of unsalted butter, cut 
        into small pieces, and mix until the 
        mixture is crumbly.
\textcolor{teal}{\textbf{[prev_steps]}}: 1. Preheat oven to 375°F (190°C). 
              2. In a large mixing bowl, combine 
              2 cups of bread flour, 1 tsp salt,
              and 1 tsp of garlic powder.
\textbf{Description: ....}
\end{coloredlcverbatim}

\noindent\textbf{\planscore.} We utilize the following prompt template for computing the \planscore:
\vspace{-0.5cm}
\begin{coloredlcverbatim}
You are my assistant in evaluating the alignment 
between the overall goal \textcolor{purple}{\textbf{[goal]}} and the 
step-by-step instructions \textcolor{blue}{\textbf{[steps]}}. 
\textcolor{purple}{\textbf{[goal]}}: How to make tomato chutney?
\textcolor{blue}{\textbf{[steps]}}: 1. Gather Ingredients ..........
Evaluate how well \textcolor{purple}{\textbf{[goal]}} aligns with \textcolor{blue}{\textbf{[steps]}}
Give a score from 0 to 100, according to  
the following criteria:
100:\textcolor{blue}{\textbf{[steps]}} perfectly describe the steps for 
    completing \textcolor{purple}{\textbf{[goal]}}.
80: \textcolor{blue}{\textbf{[steps]}} mostly describe the steps for 
    completing \textcolor{purple}{\textbf{[goal]}} but with minor 
    discrepancies in the step ordering.
60: \textcolor{blue}{\textbf{[steps]}} describe the steps for completing  
    \textcolor{purple}{\textbf{[goal]}}, but missed some important steps.
40: \textcolor{blue}{\textbf{[steps]}} didn't describe steps for completing
    \textcolor{purple}{\textbf{[goal]}} as it has discrepancies in step 
    ordering and missed few important steps.
20: \textcolor{blue}{\textbf{[steps]}} completely failed to describe the 
    steps for completing the \textcolor{purple}{\textbf{[goal]}} as it has 
    lots of discrepancies in step ordering and
    missed a lot of important steps.
Provide your analysis and explanation in JSON 
format with the following keys: 
score, explanation.  
\end{coloredlcverbatim}

The LLM returns a JSON-formatted output with \enquote{score} and \enquote{explanation} as keys. 
\vspace{0.1cm}

\noindent\textbf{\imagetotextrelevance:} For \imagetotextrelevance, we prompt VLM with two questions sequentially. First, we prompt the model to describe the contents of the visual step. 
\vspace{-0.5cm}
\begin{coloredlcverbatim}
You are my assistant to evaluate the corres-
pondence of the image to a given text prompt. 
Briefly describe the image within 50 words. 
Focus on the objects in the image and their 
attributes, such as color, shape, texture, and 
action relationships.         
\end{coloredlcverbatim}
Then, based on the answer, we ask the model to assign a step image-text cross-modal alignment score using the following prompt: 
\vspace{-0.5cm}

\begin{coloredlcverbatim}
According to the image and your previous answer, 
evaluate how well the image describes action in 
step: \textcolor{orange}{\textbf{[step]}}, a subprocess of the task \textcolor{purple}{\textbf{[goal]}}.
Give a score from 0 to 100, according to
the following criteria:
100: the image perfectly describes the action of 
     \textcolor{orange}{\textbf{[step]}} and object states after \textcolor{orange}{\textbf{[step]}}, 
     with no discrepancies.
80:  the image portrayed most of the action of 
     \textcolor{orange}{\textbf{[step]}} and object states after \textcolor{orange}{\textbf{[step]}}, 
     but with minor discrepancies.
60:  the image depicted some action of \textcolor{orange}{\textbf{[step]}} 
     and object states after \textcolor{orange}{\textbf{[step]}} but ignored 
     some key parts or details.
40:  the image did not depict any action of 
     \textcolor{orange}{\textbf{[step]}} and object states after \textcolor{orange}{\textbf{[step]}}.
20:  the image failed to convey the full action 
     of \textcolor{orange}{\textbf{[step]}} and object states after \textcolor{orange}{\textbf{[step]}}.
Provide your analysis and explanation in JSON 
format with the following keys: 
score and explanation.
\end{coloredlcverbatim}

\section{Evaluation Metrics}\label{sec:ev_metrics}
\noindent \textbf{Cross-modal Step-level Evaluation.}
We utilize {CLIPScore}~\citep{hessel-etal-2021-clipscore}, \ie CLIP embedding similarity between visual and textual steps, and our proposed {\imagetotextrelevance} that accounts for alignment in object states and implied actions. CLIPScore was originally designed for image captioning and hence may underperform in MPP settings where visual steps contain implicit cues or elements not explicitly stated in the corresponding text.
We report average CLIPScore and \imagetotextrelevance across all steps and tasks. Both scores are normalized to a 1–100 scale, with higher values indicating stronger cross-modal alignment.

\noindent \textbf{Textual Evaluation.}
In addition to \planscore, which evaluates planning accuracy and temporal coherence, we report traditional text generation metrics: \textbf{Sentence-BERT (S-BERT)}~\citep{reimers2019sentence}, \textbf{Word Movers Distance (WMD)}~\citep{kusner2015word}, and \textbf{METEOR}~\citep{banerjee-lavie-2005-meteor}. S-BERT and WMD measure feature-level similarity, and METEOR captures word-level lexical similarity between generated and reference text plans. Following TIP, we compute WMD-based similarity over sentence embeddings, where higher values denote stronger alignment. All metrics are reference-based and normalized to [0,1], whereas \planscore produces reference-free scores in [0,100].

\noindent \textbf{\visualsequenceordering.}
\visualsequenceordering evaluates predicted step order with position-based metrics. \textbf{Accuracy (Acc)} is the percentage of steps in the correct absolute position (range 0–100), and \textbf{Distance (Dist)} is the average positional deviation (Dist $\geq 0$). \textbf{Longest Common Subsequence (LCS)} measures the average overlap in subsequences (0 to sequence length), while \textbf{Kendall’s Tau ($\bm{\tau}$)}~\citep{lapata2003probabilistic} quantifies pairwise order consistency via $\tau\!=\!1-\frac{2~*~\#inversion}{\#pairs}$, where $\#inversion$ is the number of pairs in the predicted order with incorrect relative order, and $\#pairs\!=\!\binom n2$, with $\tau$ ranging from –1 to 1. \textbf{Minimum Swap (MS)} is the minimum number of swaps needed to recover the correct order (0 to sequence length–1), and \textbf{Weighted Minimum Swap (WMS)} penalizes larger swap distances (non-negative, unbounded). Higher Acc, LCS, and $\tau$ indicate stronger ordering, while lower Dist, MS, and WMS indicate fewer deviations. Following~\citet{wu2022understanding}, we evaluate on the first five sequence steps.

\section{Ablations on \modelnamenc Components}\label{sup:ablplanner}
\subsection{OSR-CoT Prompt Ablation}\label{subsec:osrcot_prompt_ablation}
OSR-CoT consists of three key components: (1) a one-shot example illustrating the image description generation process (\textbf{1-Shot}), (2) reasoning about the current step (\textbf{Desc.}), and (3) reasoning about object state changes before and after the step (\textbf{State}). 
To assess the contribution of each component, we conduct an ablation study with three variants, where components are added incrementally. In OSR-CoT-V1, the LLM is prompted to generate an image description using only the \textcolor{purple}{[goal]}, \textcolor{orange}{[step]}, and \textcolor{teal}{[prev\_steps]} without utilizing any of these components. OSR-CoT-V2 adds the one-shot example to guide the model with a concrete reference. The detailed prompt for OSR-CoT-V2 is as follows:
\vspace{-0.4cm}
\begin{coloredlcverbatim}
Write description of the \textcolor{orange}{\textbf{[step]}} image
containing the 3 sentences after \textcolor{orange}{\textbf{[step]}}.
Use \textcolor{teal}{\textbf{[prev_steps]}} for background information.
\textcolor{maroon}{\textbf{[goal]:} Task: How to make fried egg with cheese.}
\textcolor{maroon}{\textbf{[step]:} 3. Pour a small amount of butter} 
\textcolor{maroon}{        or oil into the pan.}
\textcolor{maroon}{\textbf{[prev_steps]:} 1. Crack an egg into a bowl.} 
\textcolor{maroon}{              2. Heat a non-stick frying pan} 
\textcolor{maroon}{              on medium heat.}
\textcolor{maroon}{\textbf{Image Description:}} \textcolor{maroon}{A non-stick frying pan with}
\textcolor{maroon}{butter or oil poured into it.} 

\textcolor{purple}{\textbf{[goal]}}: How to make a cheesy garlic pull-apart 
        bread?
\textcolor{orange}{\textbf{[step]}}: 3. Add 1 cup of unsalted butter, cut 
        into small pieces, and mix until the 
        mixture is crumbly.
\textcolor{teal}{\textbf{[prev_steps]}}: 1. Preheat oven to 375°F (190°C). 
              2. In a large mixing bowl, combine 
              2 cups of bread flour, 1 tsp salt,
              and 1 tsp of garlic powder.
\textbf{Image Description: ....}
\end{coloredlcverbatim}

\begin{table*}[t!]
\centering
\resizebox{0.99\linewidth}{!}{%
\begin{tabular}{c l ccc cccccccc}
\toprule
\textbf{Dataset}& \textbf{Model} &  \textbf{1-Shot} &  \textbf{Desc.} &  \textbf{State} &  CA-Score $\uparrow$ & CLIPScore $\uparrow$ & Acc $\uparrow$ & LCS $\uparrow$ & $\tau$ $\uparrow$ & Dist.$\downarrow$ & MS $\downarrow$ & WMS $\downarrow$ \\
    \hline
    \multirow{4}{*}{\rotatebox[origin=c]{90}{{\textbf{\textsc{Recipe}}}}}  \multirow{4}{*}{\rotatebox[origin=c]{90}{{\textbf{\textsc{Plan}}}}} 
    & OSR-CoT-V1  & \color{red} \ding{55} & \color{red} \ding{55} & \color{red} \ding{55}& 71.85 & 75.11 & 24.42 & 2.88 & 0.07 & 7.59 & 2.67 & 5.87   \\
    & OSR-CoT-V2  & \color{green}\textbf{\ding{51}} & \color{red} \ding{55} & \color{red} \ding{55}& 73.42 & 75.62 & 24.49 & 2.90 & 0.14 & 6.99 & 2.68 & 5.59 \\
    & OSR-CoT-V3  & \color{green}\textbf{\ding{51}} & \color{green}\textbf{\ding{51}} & \color{red} \ding{55} & 74.33 & 77.18 & 26.21 & 3.01 & 0.16 & 6.82 & 2.48 & 5.13 \\
    &  \cellcolor{lightgray!30}OSR-CoT (\modelname) & \cellcolor{lightgray!30}\color{green}\textbf{\ding{51}} & \cellcolor{lightgray!30}\color{green}\textbf{\ding{51}} & \cellcolor{lightgray!30}\color{green}\textbf{\ding{51}}
    & \cellcolor{lightgray!30}\textbf{77.07} & \cellcolor{lightgray!30}\textbf{77.55} &\cellcolor{lightgray!30}\textbf{27.50}  &\cellcolor{lightgray!30}\textbf{3.09} &\cellcolor{lightgray!30}\textbf{0.22} &\cellcolor{lightgray!30}\textbf{6.51} &\cellcolor{lightgray!30}\textbf{2.39} &\cellcolor{lightgray!30}\textbf{4.99} \\
    \hline
    \multirow{4}{*}{\rotatebox[origin=c]{90}{{\textbf{\textsc{Wiki}}}}}  \multirow{4}{*}{\rotatebox[origin=c]{90}{{\textbf{\textsc{Plan}}}}} 
    & OSR-CoT-V1 & \color{red} \ding{55} & \color{red} \ding{55} & \color{red} \ding{55} & 70.31 & 72.98 & 20.10&2.80 &0.02 &7.90 &2.70 &6.30   \\
    & OSR-CoT-V2  & \color{green}\textbf{\ding{51}} & \color{red} \ding{55} & \color{red} \ding{55} & 72.11 & 74.67 & 23.17 & \textbf{2.98} & \textbf{0.05} & 7.67 & 2.66 & 5.97 \\
    & OSR-CoT-V3  & \color{green}\textbf{\ding{51}} & \color{green}\textbf{\ding{51}} & \color{red} \ding{55} & 75.42 & 75.59 & \textbf{23.72} & 2.82 & 0.04 & 7.71 & 2.69 & \textbf{5.87} \\
    & \cellcolor{lightgray!30}OSR-CoT (\modelname)  & \cellcolor{lightgray!30}\color{green}\textbf{\ding{51}} & \cellcolor{lightgray!30}\color{green}\textbf{\ding{51}} & \cellcolor{lightgray!30}\color{green}\textbf{\ding{51}} & \cellcolor{lightgray!30}\textbf{77.44} & \cellcolor{lightgray!30}\textbf{76.10} &\cellcolor{lightgray!30}23.43 &\cellcolor{lightgray!30}2.91 &\cellcolor{lightgray!30}\textbf{0.05} &\cellcolor{lightgray!30}\textbf{7.60} &\cellcolor{lightgray!30}\textbf{2.60} &\cellcolor{lightgray!30}5.90 \\
    
    \bottomrule
\end{tabular}
}
\vspace{-0.3cm}
\caption{\textbf{Ablation on OSR-CoT Components.} Incrementally adding each component improves cross-modal alignment and temporal coherence in the generated visual plans.}
\label{tab:ablation_studies_osr_cot}
\end{table*}

\begin{table*}[t!]
\centering
\includegraphics[width=\linewidth]{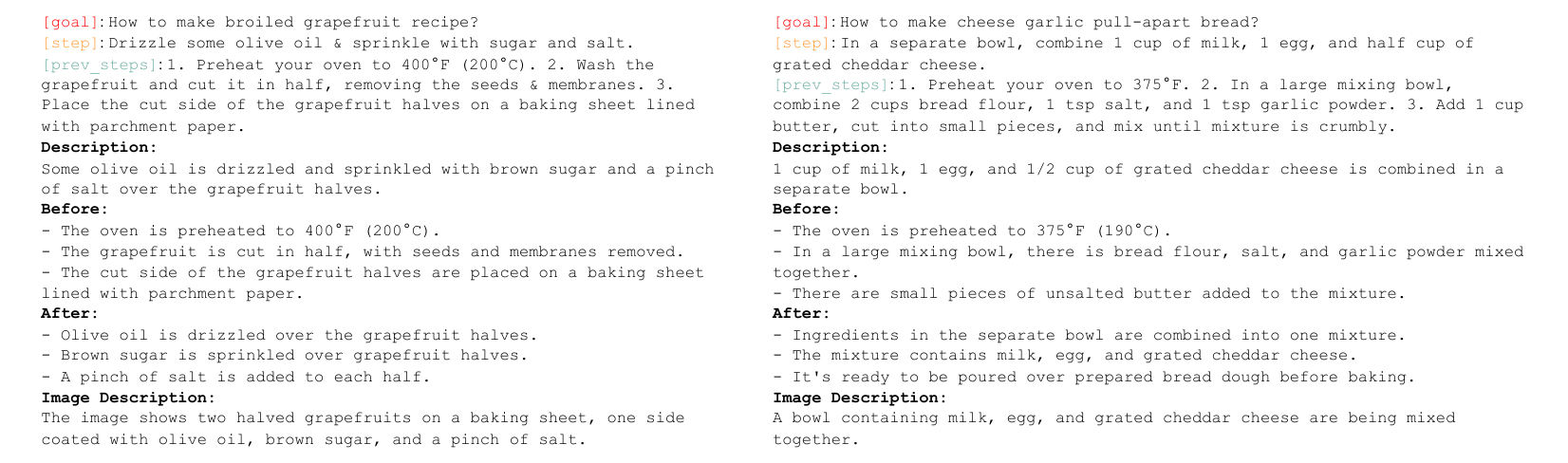}
\vspace{-0.7cm}
\caption{\textbf{Qualitative Examples} of generated descriptions by prompting with the proposed OSR-CoT method.}
\label{tab:image_description_example}
\vspace{-0.3cm}
\end{table*}

OSR-CoT-V3 extends V2 by incorporating the description component, prompting the model to first describe the current step in detail before generating the corresponding image description.
\vspace{-0.4cm}
\begin{coloredlcverbatim}
First, describe details of \textcolor{orange}{\textbf{[step]}} for \textcolor{purple}{\textbf{[goal]}}
with one verb. Use \textcolor{teal}{\textbf{[prev_steps]}} for 
background information.
Second, write description of the \textcolor{orange}{\textbf{[step]}} image
containing the 3 sentences after \textcolor{orange}{\textbf{[step]}}.
\textcolor{maroon}{\textbf{[goal]:} Task: How to make fried egg with cheese.}
\textcolor{maroon}{\textbf{[step]:} 3. Pour a small amount of butter} 
\textcolor{maroon}{        or oil into the pan.}
\textcolor{maroon}{\textbf{[prev_steps]:} 1. Crack an egg into a bowl.} 
\textcolor{maroon}{              2. Heat a non-stick frying pan} 
\textcolor{maroon}{              on medium heat.}
\textcolor{maroon}{\textbf{Description:}} 
\textcolor{maroon}{Pour a small amount of butter or oil into a pan.} 
\textcolor{maroon}{\textbf{Image Description:}}
\textcolor{maroon}{A non-stick frying pan with butter or oil}
\textcolor{maroon}{poured into it.} 
\textcolor{purple}{\textbf{[goal]}}: How to make a cheesy garlic pull-apart 
        bread?
\textcolor{orange}{\textbf{[step]}}: 3. Add 1 cup of unsalted butter, cut 
        into small pieces, and mix until the 
        mixture is crumbly.
\textcolor{teal}{\textbf{[prev_steps]}}: 1. Preheat oven to 375°F (190°C). 
              2. In a large mixing bowl, combine 
              2 cups of bread flour, 1 tsp salt,
              and 1 tsp of garlic powder.
\textbf{Description: ....}
\end{coloredlcverbatim}
Finally, the full OSR-CoT prompt incorporates the state-change component, guiding the model to reason about object transitions before and after each step. The complete prompt is provided in Appendix~\ref{sup:implementation_detail}. As shown in Table~\ref{tab:ablation_studies_osr_cot}, both \imagetotextrelevance and CLIPScore improve steadily with the inclusion of each component, underscoring their collective role in producing accurate, state-aware image descriptions. Table~\ref{tab:image_description_example} presents qualitative examples.

\begin{table}[t!]
\centering
\renewcommand{\arraystretch}{1.3}
\resizebox{\columnwidth}{!}{%
\begin{tabular}{clcccccc}
\toprule
\textbf{Dataset} & \textbf{FE} & \imagetotextrelevance $\uparrow$& CLIPScore   
 $\uparrow$ & $\tau$ $\uparrow$& Dist.$\downarrow$ & MS $\downarrow$ & WMS $\downarrow$ \\
\midrule
\multirow{3}{*}{\rotatebox[origin=c]{90}{{\textbf{\textsc{Recipe}}}}}  \multirow{3}{*}{\rotatebox[origin=c]{90}{{\textbf{\textsc{Plan}}}}} 
& No FE & 75.27 & 75.19 & 0.16 &6.82 &2.48 &5.03 \\
& CLIP  & 75.85 & 75.67 & 0.14 &7.01 &2.46 &5.15 \\
&  \cellcolor{lightgray!30}BLIP-2& \cellcolor{lightgray!30}\textbf{77.07} & \cellcolor{lightgray!30}\textbf{77.55} &\cellcolor{lightgray!30}\textbf{0.22} &\cellcolor{lightgray!30}\textbf{6.51} &\cellcolor{lightgray!30}\textbf{2.39} &\cellcolor{lightgray!30}\textbf{4.99} \\
\cmidrule(lr){1-8}
\multirow{3}{*}{\rotatebox[origin=c]{90}{{\textbf{\textsc{Wiki}}}}}  \multirow{3}{*}{\rotatebox[origin=c]{90}{{\textbf{\textsc{Plan}}}}} 
& No FE & 75.70 & 74.35 &0.03 &7.78 &2.60 &6.16 \\
& CLIP   & 75.76 & 76.00 & 0.04 &7.67 &2.64 &5.97\\
& \cellcolor{lightgray!30}BLIP-2& \cellcolor{lightgray!30} \textbf{77.44} & \cellcolor{lightgray!30}\textbf{76.10} &\cellcolor{lightgray!30}\textbf{0.05} &\cellcolor{lightgray!30}\textbf{7.60} &\cellcolor{lightgray!30}\textbf{2.60} &\cellcolor{lightgray!30}\textbf{5.90} \\
\bottomrule
\end{tabular}
}
\vspace{-0.3cm}
\caption{\textbf{Ablation on Cross-Modal Feature Extractors (FEs)} with no feature extractor ({No FE}), CLIP, and BLIP-2 for cross-modal step image selection.} 
\label{tab:ablation_studies_feature_extractor}
\vspace{-0.3cm}
\end{table}

\subsection{Cross-modal Feature Extractor}\label{subsec:crossmodal_feature_extractor}
\textbf{Effectiveness of BLIP-2 as Feature Extractor.} 
To evaluate the impact of different cross-modal feature extractors (FEs) in step image selection (Section~\ref{subsec:visual_plan_generation}), we compare: (1) \textbf{No FE}, which selects an image based solely on the step image description without any kind of cross-modal feature extraction; (2) \textbf{CLIP}, a retrieval-based model; and (3) \textbf{BLIP-2}, the cross-modal feature extractor used in \modelnamenc. As shown in Table~\ref{tab:ablation_studies_feature_extractor}, BLIP-2 consistently outperforms other variants, likely due to its alignment strategies, which better capture fine-grained visual-textual correspondence.
\begin{table}[t!]
\centering
\resizebox{\columnwidth}{!}{%
\begin{tabular}{c l cccccc}
\toprule
\textbf{Dataset}& \textbf{\textit{K}} & CA-Score $\uparrow$ & CLIPScore $\uparrow$ & $\tau$ $\uparrow$ & Dist.$\downarrow$ & MS $\downarrow$ & WMS $\downarrow$ \\
    \midrule
    \multirow{5}{*}{\rotatebox[origin=c]{90}{{\textbf{\textsc{Recipe}}}}}  \multirow{5}{*}{\rotatebox[origin=c]{90}{{\textbf{\textsc{Plan}}}}} 
    & 1  &75.27 &75.19 &0.16 &6.82 &2.48 &5.02 \\
    & 5  &75.34	& 75.30 &0.19 &6.76 &2.44 &\textbf{4.93} \\
    
    & 10  &75.43 & 75.27 &0.20 &6.68 &2.41 &4.97 \\
    & 15  &76.05 & 77.24 &0.20 &6.64 &2.41 &5.00 \\
    &  \cellcolor{lightgray!30}20 &\cellcolor{lightgray!30}\textbf{77.07}  &\cellcolor{lightgray!30}\textbf{77.55} &\cellcolor{lightgray!30}\textbf{0.22} &\cellcolor{lightgray!30}\textbf{6.51} &\cellcolor{lightgray!30}\textbf{2.39} &\cellcolor{lightgray!30}{4.99} \\
    \cmidrule(lr){1-8}\
    
   \multirow{5}{*}{\rotatebox[origin=c]{90}{{\textbf{\textsc{Wiki}}}}}  \multirow{5}{*}{\rotatebox[origin=c]{90}{{\textbf{\textsc{Plan}}}}} 
    & 1  &75.70 &74.35 &0.03 &7.78 &2.61 &6.16 \\
    & 5  &75.94 &75.14 &0.03 &7.75 &2.63 &6.16 \\
    & 10  &75.88 &75.21 &0.04 &7.68 &2.64 &6.13 \\
    & 15  &76.11 &76.12 &0.04 &7.71 &2.61 &6.13 \\
    &  \cellcolor{lightgray!30}20  &\cellcolor{lightgray!30}\textbf{77.44}  &\cellcolor{lightgray!30}\textbf{76.10} &\cellcolor{lightgray!30}\textbf{0.05} &\cellcolor{lightgray!30}\textbf{7.60} &\cellcolor{lightgray!30}\textbf{2.60} &\cellcolor{lightgray!30}\textbf{5.90} \\
    
    \bottomrule
\end{tabular}
}
\vspace{-0.3cm}
\caption{\textbf{Ablation on Cross-Modal Step Image Selection Hyperparameter \textit{K}} (number of generated images).}
\label{tab:ablation_studies_sensitivity_k}
\end{table}

\noindent\textbf{Ablation on Hyperparameter $K$.} 
We investigate the effect of the hyperparameter $K$ in the cross-modal step image selector, which determines the number of candidate images generated per step. Specifically, we vary $K$ across the following values: $K\in\{1, 5, 10, 15, 20\}$. As shown in Table~\ref{tab:ablation_studies_sensitivity_k}, increasing $K$ leads to consistent gains in \imagetotextrelevance and CLIPScore, indicating improved alignment between selected images and step text. While improvements in \visualsequenceordering metrics are modest, the results suggest that higher $K$ values enhance visual relevance and semantic fidelity w.r.t. the corresponding step texts.

\subsection{Ablation on Multimodal Goals}\label{subsec:ablation_multimodal_goal}
We conduct an ablation to analyze the role of goal image $\mathcal{G}_v$ in generating textual plans. As shown in Table~\ref{tab:visual_goal_ablation}, incorporating the visual goal consistently improves performance compared to using the textual goal alone, demonstrating that the goal image $\mathcal{G}_v$ contributes complementary information that enhances the quality of generated textual plans. 

\begin{table}[t!]
\centering
\renewcommand{\arraystretch}{1.1}
\resizebox{\columnwidth}{!}{%
\begin{tabular}{clcccc}
\toprule
\textbf{Dataset} & $\boldsymbol{\mathcal{G}}_v$ & \planscore $\uparrow$ & S-BERT $\uparrow$ & WMD $\uparrow$ & METEOR $\uparrow$ \\
\midrule
    \multirow{2}{*}{\rotatebox[origin=c]{90}{\small {\textbf{\textsc{Recipe}}}}}  \multirow{2}{*}{\rotatebox[origin=c]{90}{{\small \textbf{\textsc{Plan}}}}} & \color{red}\ding{55} & 80.88 & 0.75 & 0.76 & 0.10 \\
    & \cellcolor{lightgray!30}\color{green}\textbf{\ding{51}} & \cellcolor{lightgray!30}\textbf{82.05} & \cellcolor{lightgray!30}\textbf{0.78} & \cellcolor{lightgray!30}\textbf{0.88} & \cellcolor{lightgray!30}\textbf{0.15} \\
\cmidrule(lr){1-6}
     \multirow{2}{*}{\rotatebox[origin=c]{90}{{\small \textbf{\textsc{Wiki}}}}}  \multirow{2}{*}{\rotatebox[origin=c]{90}{{\small \textbf{\textsc{Plan}}}}} 
     &  \color{red}\ding{55} & 82.30 & \textbf{0.77} & \textbf{0.76} & 0.20 \\
     & \cellcolor{lightgray!30}\color{green}\textbf{\ding{51}} & \cellcolor{lightgray!30}\textbf{84.43} & \cellcolor{lightgray!30}\textbf{0.77} & \cellcolor{lightgray!30}0.75 & \cellcolor{lightgray!30}\textbf{0.23} \\
\bottomrule
\end{tabular}
}
\vspace{-0.2cm}
\caption{\textbf{Ablation on Multimodal Goal.} Generating textual plans w/ and w/o a Goal Image.} 
\label{tab:visual_goal_ablation}
\vspace{-0.3cm}
\end{table}

 \begin{table*}[t!]
\centering
\resizebox{0.9\textwidth}{!}{%
\begin{tabular}{lcccccccc}
\toprule
\multirow{3}{*}{\textbf{Model}} & \multicolumn{4}{c}{\textbf{\imagetotextrelevance}} & \multicolumn{4}{c}{\textbf{\planscore}}\\
& \multicolumn{2}{c}{\textsc{\textbf{RecipePlan}}} & \multicolumn{2}{c}{\textsc{\textbf{Wikiplan}}} & \multicolumn{2}{c}{\textsc{\textbf{RecipePlan}}} & \multicolumn{2}{c}{\textsc{\textbf{Wikiplan}}} \\
\cmidrule(lr){2-5}\cmidrule(lr){6-9}
& MiniGPT-4 & LLaVa & MiniGPT-4 & LLaVa & GPT-3.5 & LLaMa & GPT-3.5 & LLaMa\\ 
    \midrule
    Text-Ref + SD & 71.47 & 55.65 & 70.81 &54.51 & 73.85 & 71.66 & 62.53 & 55.20 \\
    GPT-3.5 + SD & 72.24& 56.42 & 71.49& 55.52 & 80.43 & 85.86 & 81.93 & 85.36\\
    TIP & 70.08	& 54.49 & 67.68 & 55.76 & 82.00 & 87.12 & 83.00 & 90.43\\
     \rowcolor{lightgray!30} \textbf{\modelname} & \textbf{77.07} & \textbf{63.10} &\textbf{77.44} &\textbf{62.59}  &\textbf{82.05} & \textbf{87.67} & \textbf{84.43} & \textbf{90.46}\\
    \bottomrule
\end{tabular}
}
\vspace{-0.3cm}
\caption{\textbf{Ablation on LLM/VLMs} used in LM-based evaluation  \imagetotextrelevance and \planscore.}
\label{tab:ablation_ca_tplan_score}
\end{table*}

\begin{table*}[t!]
\centering
\resizebox{0.99\textwidth}{!}{%
\begin{tabular}{cccccccccccc}
\toprule
\multirow{2}{*}{ \textbf{Permute}} & \multirow{2}{*}{ \textbf{Delete}} & \multicolumn{5}{c}{\textsc{ \textbf{RecipePlan}}} & \multicolumn{5}{c}{\textsc{ \textbf{Wikiplan}}} \\
\cmidrule(lr){3-7}\cmidrule(lr){8-12} 
&& \planscore (GPT-3.5) & \planscore (LLaMa) & WMD & METEOR & S-BERT & \planscore (GPT-3.5) & \planscore (LLaMa) & WMD & METEOR & S-BERT \\ 
    \midrule
    {\color{green}\textbf{\ding{51}}} &  {\color{green}\textbf{\ding{51}}} & 77.50 & 55.8 & 0.87 & 0.09 & 0.76 & 78.10 & 66.34 & 0.74 & 0.19 & \textbf{0.77}\\
     {\color{red} \ding{55}} &  {\color{green}\textbf{\ding{51}}} & 78.12 & 60.68 & 0.87 & 0.09 & 0.76 & 80.10 &70.75 &  	0.74 & 0.19 & 0.76\\
      {\color{green}\textbf{\ding{51}}} & {\color{red} \ding{55}} & 78.60	& 64.78 & \textbf{0.89} & \textbf{0.15} & \textbf{0.77} & 80.20& 74.64 & \textbf{0.75} & \textbf{0.23} & \textbf{0.77} \\
    {\color{red} \ding{55}} & {\color{red} \ding{55}} & \textbf{82.05} & \textbf{79.81} &  	0.88 &	\textbf{0.15} &	\textbf{0.77} &\textbf{84.43} &\textbf{86.07} &  	\textbf{0.75} &	\textbf{0.23} &	\textbf{0.77}\\
    \bottomrule
\end{tabular}
}
\vspace{-0.3cm}
\caption{\textbf{Verifying \planscore} on \modelnamenc's textual plans with unordered or missing steps.}
\label{tab:ablation_correctness_tplan}
\end{table*}

\begin{table}[t!]
\centering
\resizebox{0.9\columnwidth}{!}{%
\begin{tabular}{lcccc}
\toprule
\multirow{2}{*}{\textbf{Deletion \%}} &  \multicolumn{2}{c}{\textsc{ \textbf{RecipePlan}}} & \multicolumn{2}{c}{\textsc{\textbf{Wikiplan}}} \\
\cmidrule(lr){2-3}\cmidrule(lr){4-5} 
& GPT-3.5 & LLaMa & GPT-3.5 & LLaMa\\ 
    \midrule
    80  & 75.02 & 50.56 & 77.14 & 68.01\\
     60 & 76.88 & 61.23 & 79.26 &75.04\\
      50 & 78.12	& 70.13 & 80.10& 77.18\\
      40 & 78.38	& 76.02 & 80.43& 80.70\\
      20 & 80.51	& 80.10 & 81.57& 82.55\\
    0  & \textbf{82.05} & \textbf{86.67} &\textbf{84.43} &\textbf{89.46} \\
    \bottomrule
\end{tabular}
}
\vspace{-0.3cm}
\caption{\textbf{\planscore Ablation} with varying $\%$ of missing steps for plans generated by \modelnamenc.} 
\label{tab:ablation_correctness_tplan_deletion_percentage}
\vspace{-0.3cm}
\end{table}

\section{Ablations on MPP Evaluation}\label{sec:abl_metrics}
\subsection{Robustness of \planscore and \imagetotextrelevance}\label{subsec:ablation_llms_for_scores} 
We evaluate \planscore and \imagetotextrelevance across different LLMs and VLMs. As shown in Table~\ref{tab:ablation_ca_tplan_score}, \modelnamenc consistently achieves the highest scores across all configurations, demonstrating its effectiveness independent of the underlying evaluation model. Moreover, the consistent trends across baselines confirm that both \planscore and \imagetotextrelevance serve as stable and reliable metrics for evaluating procedural plans.

\subsection{\planscore Reliability}\label{subsec:abl_tplan} 
Both \textsc{RecipePlan} and \textsc{WikiPlan} include tasks requiring domain expertise, such as \enquote{\textit{How to fix a leaky faucet}} and \enquote{\textit{How to pasteurize}}. Given the complexity of these tasks, human evaluation for assessing plan accuracy would be costly and labor-intensive. 
Instead, to evaluate the reliability of \planscore, we conduct an ablation study by perturbing LLM-generated plans using two strategies: (i) \textit{random permutation}, which shuffles the step order; and (ii) \textit{random deletion}, which randomly removes $50\%$ of the textual plan steps. 
We compute \planscore using both GPT-3.5 and LLaVa-1.5-13B to assess its robustness across model types. As shown in Table~\ref{tab:ablation_correctness_tplan}, \planscore consistently degrades when steps are deleted or permuted, demonstrating its sensitivity to structural disruptions in the plan. In contrast, standard metrics such as WMD, METEOR, and S-BERT exhibit minimal variation and fail to capture these structural inconsistencies. This highlights \planscore's unique ability to penalize violations in temporal coherence, which are often overlooked by traditional text similarity metrics. Furthermore, we vary the deletion percentage to test granularity. Table~\ref{tab:ablation_correctness_tplan_deletion_percentage} shows \planscore increases with more complete plans, demonstrating its sensitivity to missing steps.

\subsection{\imagetotextrelevance and Human Correlation}\label{subsec:ca_score_correlation}
To evaluate how well \imagetotextrelevance aligns with human judgment, we conducted a human study involving 30 step-image examples: 10 from ground truth, 10 from TIP, and 10 from \modelnamenc. 14 annotators independently assessed each image's relevance to its paired textual instruction, resulting in 420 human ratings in total (30 examples × 14 raters). Evaluations were performed on a 5-point Likert scale, with annotators instructed to consider both the depicted action and object states, as well as the image’s alignment with the overarching task goal. A score of 1 reflects an image that is irrelevant to both the step and the overall goal, whereas a score of 5 signifies perfect alignment with both. 
To evaluate inter-annotator agreement, we compute both the weighted Cohen's kappa and Spearman's rank correlation coefficient~\citep{sedgwick2014spearman},  obtaining scores of 0.61 and 0.67, respectively, indicating moderately strong inter-rater agreement.

Table~\ref{tab:comparison_with_human} presents the average human rating along with \imagetotextrelevance and CLIPScore for the collected examples, suggesting that raters preferred the step images generated from \modelnamenc while step images generated by TIP are perceived to be less accurate or relevant compared to the ground truth and \modelnamenc generated plans. Finally, to assess the alignment between automated metrics and human evaluation, we compute Spearman’s rho ($\rho$) for both \imagetotextrelevance and CLIPScore against human ratings, yielding a correlation of $0.57$ for \imagetotextrelevance and $0.37$ for CLIPScore, suggesting \imagetotextrelevance reflects human judgment better.

\subsection{Motivation of \visualsequenceordering Task}\label{sec:multseqord}
\citet{wu2022understanding} propose a multimodal sequencing task that assesses temporal coherence by predicting the correct order of an unordered multimodal plan (text and image steps). We apply this task to evaluate the output of baseline models under the hypothesis that more interpretable and expressive plans would yield higher reordering accuracy. However, as shown in Table~\ref{tab:multimodal_vs_visual_sequence_ordering}, all baselines perform similarly, largely due to their accurate textual plans. Since the sequencing model primarily relies on textual cues, improvements in visual quality have a limited impact. To better isolate visual coherence, we instead adopt the vision-only reordering model from~\citet{wu2022understanding}, where gains in visual planning directly enhance task performance.

\begin{table}[t!]
\centering
\resizebox{0.99\columnwidth}{!}{%
\begin{tabular}{cccc}
\toprule
\textbf{Model} & \imagetotextrelevance $\uparrow$& CLIPScore   
 $\uparrow$ & Human Rating $\uparrow$ \\
\midrule
\textcolor{battleshipgrey}{GroundTruth} & \textcolor{battleshipgrey}{86.00} & \textcolor{battleshipgrey}{77.16} & \textcolor{battleshipgrey}{4.19} \\
TIP  & 58.00 & 76.00 & 2.51 \\
\cellcolor{lightgray!30} \textbf{\modelname} & \cellcolor{lightgray!30}\textbf{85.00} & \cellcolor{lightgray!30}\textbf{79.19} &\cellcolor{lightgray!30}\textbf{4.19} \\
\bottomrule
\end{tabular}
}
\vspace{-0.3cm}
\caption{\textbf{Comparison of \imagetotextrelevance, CLIPScore, and Human Ratings} for step image-text pairs evaluated by humans across ground truth (GroundTruth), TIP, and \modelnamenc generated plans.} 
\label{tab:comparison_with_human}
\end{table}

\begin{table}[t!]
\centering
\renewcommand{\arraystretch}{1.1} 
\resizebox{0.99\columnwidth}{!}{%
\begin{tabular}{clcccccc}
\toprule
\textbf{Modality} & \textbf{Model} & Acc $\uparrow$ & LCS $\uparrow$ & $\tau$ $\uparrow$& Dist.$\downarrow$ & MS $\downarrow$ & WMS $\downarrow$ \\
    \midrule
    \multirow{2}{*}{\rotatebox[origin=c]{90}{{\textbf{Multi}}}}  \multirow{2}{*}{\rotatebox[origin=c]{90}{{\textbf{modal}}}} 
    & TIP   &74.20 &4.34 &0.80 &1.80 &0.75 &1.02 \\
    & \cellcolor{lightgray!30}\textbf{\modelname} & \cellcolor{lightgray!30}\textbf{74.43}  &\cellcolor{lightgray!30}\textbf{4.37} &\cellcolor{lightgray!30}\textbf{0.80} &\cellcolor{lightgray!30}\textbf{1.70} &\cellcolor{lightgray!30}\textbf{0.73} &\cellcolor{lightgray!30}\textbf{1.01}\\
    \cmidrule(lr){1-8}
    
    \multirow{2}{*}{\rotatebox[origin=c]{90}{{\textbf{Vision}}}}  \multirow{2}{*}{\rotatebox[origin=c]{90}{{\textbf{Only}}}} 
    
    & TIP   &21.70 &1.81 &0.05 &7.79 &2.66 &5.99 \\
    & \cellcolor{lightgray!30}\textbf{\modelname} & \cellcolor{lightgray!30}\textbf{27.50}  &\cellcolor{lightgray!30}\textbf{3.09} &\cellcolor{lightgray!30}\textbf{0.22} &\cellcolor{lightgray!30}\textbf{6.51} &\cellcolor{lightgray!30}\textbf{2.39} &\cellcolor{lightgray!30}\textbf{4.99}\\
    
    \bottomrule
\end{tabular}
}
\vspace{-0.3cm}
 \caption{
    \textbf{Comparison of Multimodal and Vision-Only Sequence Ordering on \textsc{RecipePlan}.}}
    \label{tab:multimodal_vs_visual_sequence_ordering}
    \vspace{-0.3cm}
\end{table}

\section{Qualitative Examples}\label{sec:qualexamples}
\begin{figure*}[t!]
\centering
  \includegraphics[width=0.99\textwidth]{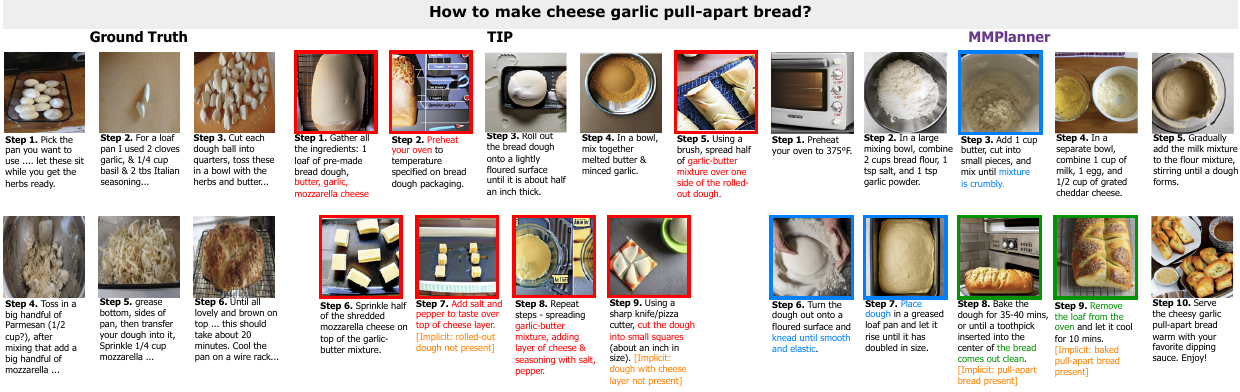}
   \vspace{-0.4cm}
  \caption{\textbf{Qualitative Comparison between TIP and \modelnamenc on \textsc{RecipePlan} for the task goal \enquote{How to make cheese garlic pull-apart bread}.} \underline{Left:} In step 7, TIP fails to incorporate the dough state in the generated step image, as it was not mentioned in the textual step (\textcolor{orange}{orange text}). Moreover, in step 6, TIP does not depict the \enquote{shredded cheese} in the step image, which is explicitly mentioned in the textual step (\textcolor{red}{red text and bboxes}). \underline{Right:} In step 9, \modelnamenc depicts the correct state of \enquote{baked loaf} (\textcolor{green}{green}) even if it was not mentioned in the textual step (\textcolor{orange}{orange text}). In step 3, the generated step image illustrates the explicit object state \enquote{crumbly} (\textcolor{blue}{blue}).}
  \label{fig:qualitative_recipe_2}
  \vspace{0.1cm}
\end{figure*}

\begin{figure*}[t!]
\centering
  \includegraphics[width=0.99\textwidth]{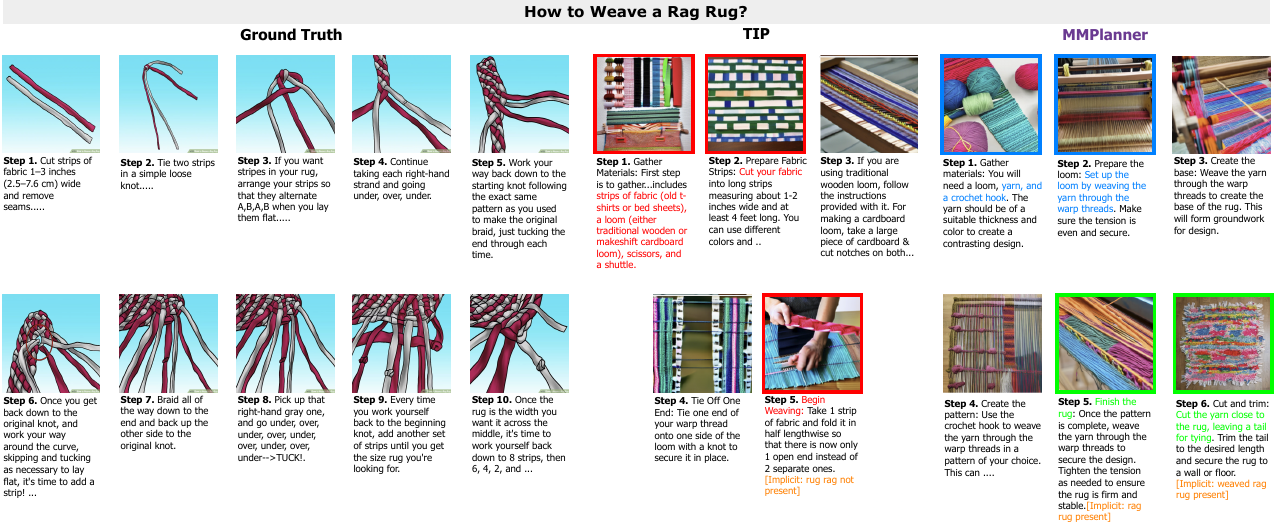}
   \vspace{-0.4cm}
  \caption{\textbf{Qualitative Comparison between TIP and \modelnamenc on \textsc{WikiPlan} for the task goal \enquote{How to Get a Sick Kitten to Eat}.} \underline{Left:} In step 1, TIP fails to incorporate the explicit object states in the generated step image (\textcolor{red}{red texts and boxes}). \underline{Right:} In step 1, \modelnamenc incorporates the explicit state of the foods (\textcolor{blue}{blue}).} 
  \label{fig:qualitative_wiki_2}
  \vspace{0.1cm}
\end{figure*}

\begin{figure*}[t!]
\centering
  \includegraphics[width=0.99\textwidth]{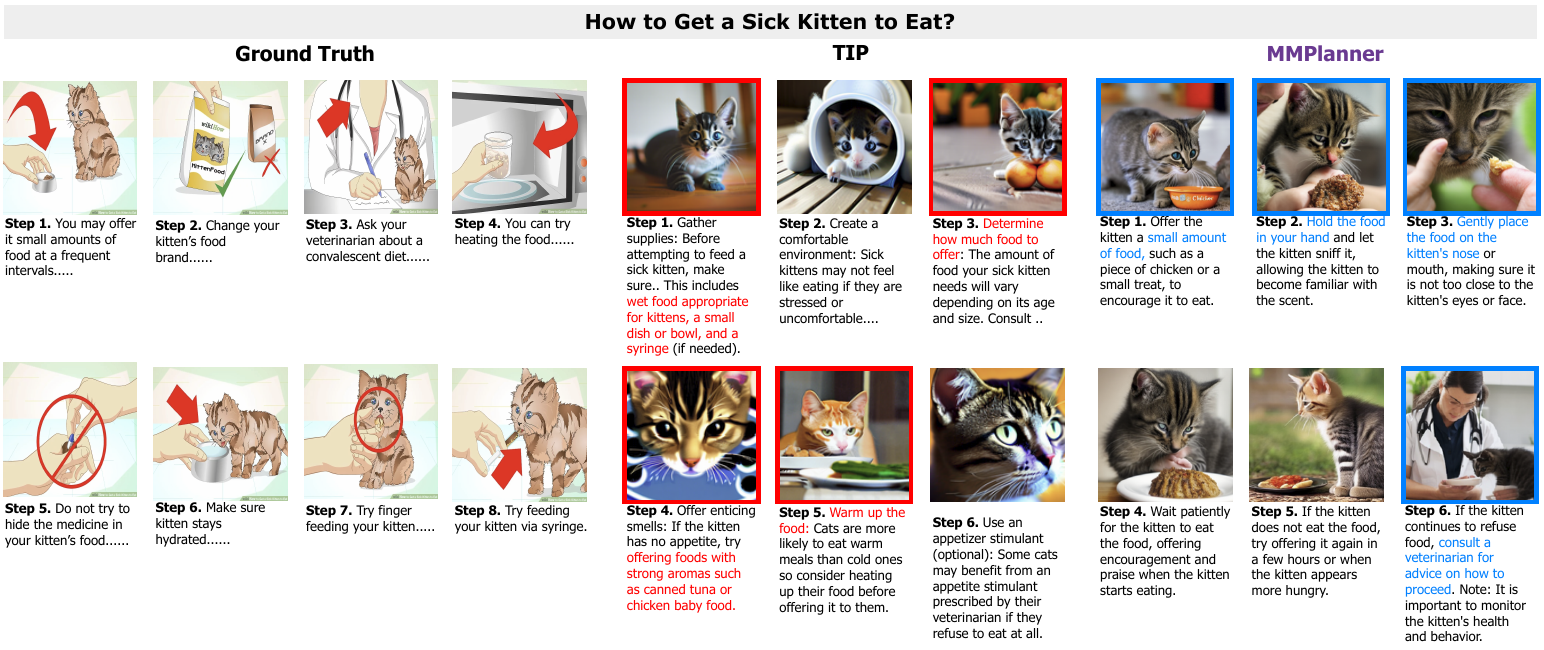}
   \vspace{-0.4cm}
  \caption{\textbf{Qualitative Comparison between TIP and \modelnamenc on \textsc{WikiPlan} for the task goal \enquote{How to weave rag rug}.} \underline{Left:} In step 1, TIP fails to incorporate the explicit object states in the generated step image (\textcolor{red}{red texts and boxes}). Moreover, in step 5, the step image does not contain the implicit object information \enquote{rag rug} (\textcolor{orange}{orange text}). \underline{Right:} In step 1, \modelnamenc incorporates the explicit state of the ingredients (\textcolor{blue}{blue}). In step 6, the generated step image includes the implicit object state \enquote{finished rag rug} (\textcolor{green}{green}).} 
  \label{fig:qualitative_wiki_1}
\end{figure*}

Figure~\ref{fig:qualitative_recipe_2} compares TIP and \modelnamenc on the task \enquote{\textit{How to make cheese garlic pull-apart bread?}}. In this example, TIP generates a generic cheese block for step 7, failing to reflect the dish-specific context (\textit{pull-apart bread}), which is not explicitly mentioned in the text. In contrast, \modelnamenc correctly depicts the baked bread in step 8, despite the step only referring to baking the dough, demonstrating its ability to infer object state transitions beyond surface text.
Figures~\ref{fig:qualitative_wiki_2} and~\ref{fig:qualitative_wiki_1} provide additional qualitative examples on two \textsc{WikiPlan} tasks, \enquote{\textit{How to Get a Sick Kitten to Eat?}} and \enquote{\textit{How to Weave a Rag Rug?}}, respectively. 
In Figure~\ref{fig:qualitative_wiki_2}, while all TIP-generated images include a kitten, they lack consistency across steps (\eg, step 5) and often omit explicit objects mentioned in the text, such as the dish/bowl in step 1 and the food in step 4. \modelnamenc produces step images that more faithfully reflect the textual instructions and maintain higher visual consistency.
In Figure~\ref{fig:qualitative_wiki_1}, TIP fails to infer implicit object state information such as \enquote{rag rug} in step 5. \modelnamenc, however, includes the rug in steps 5-6, demonstrating its ability to maintain visual consistency across step images.

\end{document}